**Neural Network Characterization and Entropy Regulated Data Balancing through Principal Component Analysis**

David Yevick and Karolina Hutchison
Department of Physics
University of Waterloo
Waterloo, ON N2L 3G7
yevick@uwaterloo.ca

**Abstract:** This paper examines in detail the geometric structure of principal component analysis (PCA) by considering in detail the distributions of both unrotated and rotated MNIST digits in the space defined by the lowest order PCA components. Since digits possessing salient geometric features are mapped to restricted regions far from the origin, they are predicted by neural networks with a greater accuracy than digits that are mapped to broad, diffuse and overlapping volumes of the low order PCA space. Motivated by these results, a new quantity, the local PCA entropy, obtained by dividing the spatial region spanned by the low order principal components into histogram bins and evaluating the entropy associated with the number of occurrences of each input class within a bin, is introduced. The metric locates the input data records that yield the largest confusion in prediction accuracy within reduced coordinate volumes that optimally discriminate among geometric features. As an example of the potential utility of the local PCA entropy, a simple data balancing procedure is realized by oversampling the data records in regions of large local entropy.

**1. Introduction:** The application of machine learning to fields such as computer vision,[1] medical imaging,[2,3] natural language processing[4] and control systems has recently led to novel results in areas such as small object detection,[5] spatiotemporal fault estimation,[6] and partial differential equation (PDE) simulation.[7] Underlying many of these advances is the ability of neural networks, when properly configured and trained, to distinguish differing classes of objects, even in the presence of irregular shape variations or geometric transformations such as translations and rotations. Unlike analytic methods, neural networks identify or classify data features through multivariable optimization. However, the accuracy of a neural network is limited if the dataset is incomplete and is further dependent on computational metaparameters such as the number and connectivity of its individual computing elements.[8] In contrast, if the underlying system is stochastic or if its mathematical description is subject to measurement uncertainty, machine learning methods can prove as accurate as deterministic procedures.

Despite the generality and simplicity of neural network methods, processing datasets with large dimensionality, typified by large images, requires substantial computational resources. As a consequence, considerable effort has been expended to develop techniques for projecting data onto a smaller number of features in order to reduce the required network complexity. For example, images arising from earth sensing of vegetation are often partly classified according to their texture[9,10] employing a variety of procedures[11] such as Markov random fields[12], granulometric analysis[13] and wavelet transforms[14]. However, the simplest, if relatively inefficient technique for analyzing the local properties of images is to perform histogram averages over the local regions delimited by sliding local window filters.[15,16] Such calculations can be accelerated by employing weighted filters[17] and characterizing the resulting histograms by statistical quantities such as their moments or the local[18] or global Shannon entropy.[19–22] Local entropy has also been defined for non-image problems such as protein chains[23] and interatomic structure[24] and has as well be extensively examined mathematically.[25–27]. While the great majority of techniques are accordingly based on local features of the input data, a method more analogous to the procedure presented in this paper applies nonlinear regression to the local entropy histograms in order to generate features that additionally characterize the global variation of this quantity.[28] However, the advantage of the PCA is that nonlinear basis functions do not have to be specified in advance while the variable combinations associated with the low order PCA basis functions yield the optimal separation among the input data classes.

To illustrate the wide potential applicability of the local PCA entropy procedure introduced below, a data balancing algorithm based on this quantity will additionally be formulated. This example addresses the dependence of the computational accuracy of a neural network calculation on the fidelity of the training data, which can be compromised by dataset imbalance if one group of classes occur far less frequently or are less easily distinguished than the remaining classes.[29–41] This leads to "minority" data records being incorrectly classified as members of the "majority" class.[30,34] For example, in education, imbalanced data sets occur when examining causes of student failure since the number of successful students typically exceeds the number of the target group by at least an order of magnitude.[35,42–44] Similarly in medicine, images of healthy tissues occur far more frequently than those of cancerous tissue [45–47], while in geology desirable minerals are typically uncommon.[48,49] Analogous issues occur in finance, ecology, telecommunications, internet programming, and biology, among other fields.[30,38,45,50–63]

Methods for balancing data sets are either algorithmic or data based.[30,34] The first of these is typified by cost-sensitive learning, which enhances the importance of the infrequent data samples of interest by multiplying the contribution of each data record in a class to the loss function by a cost that is inversely proportional to the number of samples of the class in the dataset.[64–68] Data based methods instead preprocess the input data by undersampling the majority class records or oversampling the minority records as in random undersampling and random oversampling respectively.[34,69–75] Random undersampling generates a balanced training data set by randomly excluding majority class data records, which reduces the computation time at the cost of a certain degree of information loss. In contrast, the random oversampling technique balances the data distribution by replicating minority class data records. This, however, can result in overfitting.[76]

More involved oversampling techniques often incorporate data augmentation as in the synthetic minority oversampling technique (SMOTE) [70,76–80] as well as in hybrid techniques typified by the SMOTE with edited nearest neighbors (SMOTE ENN) [81,82] and the SMOTE with Tomek links (SMOTE TOMEK).[69] The SMOTE method, which is applied to the minority members of a data set, employs the K-nearest neighbor procedure to identify the nearest neighbors of a feature vector and then to compute the distances between these vectors. This information is then employed to combine the feature vector with those of its nearest neighbors, yielding a new synthetic minority data sample.[78,83,84] SMOTE is however limited by the assumption that every minority instance possesses an identical information content. Accordingly, the SMOTE TOMEK procedure,[69] which was additionally applied to time-series data in [85] deletes data members that form Tomek links between the two classes, resulting in better defined boundaries between the class clusters. The SMOTE ENN method combines SMOTE and the edited nearest neighbor (ENN) procedure which removes data records that are misclassified relative to a prediction based on the labels of their three nearest neighbors.[81,82] This typically eliminates more samples than SMOTE TOMEK and hence yields an improved (cleaner) data set. An additional synthetic algorithm is afforded by the adaptive synthetic sampling (ADASYN) technique.[83,86,87] Data imbalance can also be addressed by adapting a learning algorithm to the imbalanced data [69] as in cost-sensitive learning techniques.[88] These are however intrinsically more complex than procedures based on data sampling.[89]

In some contexts, balancing can be further complicated by anomalies in data collection or limited processing times. For example, not only does credit card fraud occur infrequently but access to data is often restricted by privacy considerations while detection must occur rapidly in real time in order to reject fraudulent transactions.[90,91] Methods that have been applied to this problem include random sampling, which yielded the largest true positive and lowest false positive rates when a 50:50 distribution of legitimate versus non-legitimate training data was employed. Stratified undersampling of legitimate records instead exhibited optimal performance when 10% of the records were fraudulent[92] A subsequent analysis instead employed a feedforward complementary neural network (CMTNN) undersampling method. This technique eliminates redundant training samples with low information content by learning the majority class features and hence achieves a data set with a maximum information density.[93–95]

This paper examines in detail the relationship between the properties of the principal component analysis (PCA) decomposition of data at different stages in linear and nonlinear neural networks and the neural network behavior. The resulting insights motivate the novel concept of local PCA entropy, the practical utility of which is finally illustrated through a simple data balancing example.

**2. MNIST Data Set:** The geometric information contained in the low-order PCA components can be conveniently visualized in the context of the standard benchmark examples of unrotated and randomly rotated MNIST digits. These comprise a diverse collection of 70,000 handwritten digits discretized as $28 \times 28$ pixel arrays with 256 grayscale levels together with their associated labels. After normalizing these arrays so that their values fall in the interval $[0, 1]$, training and test sets are created with 60,000 and 10,000 digits, respectively. The neural network calculations below that apply PCA to these data sets are based on the easily manipulated code present in section 2.5.1 of [96].

**3. Principal Component Analysis and Neural Networks:** Principal component analysis can often be substituted for more complex elements of a machine learning algorithm without significant loss of accuracy. For example, an architecture similar to a convolutional neural network in which the learned filters were replaced by PCA eigenvectors exhibited excellent classification performance.[97] The PCA is also often employed to reduce the input data volume through preprocessing.[98,99] At the same time, however, the PCA is inefficient when applied to problems for which relevant information is embedded in a broad spectrum of high-frequency components as can occur in communication signals or visualization.[100,101]

In many cases, high dimensional data can be compressed by projecting onto the two or three dimensional space of the lowest-order PCA components. For example, Figure 1 plots the second-order PCA component against the first, lowest-order PCA component for all of the (unrotated) MNIST data records. The solid lines in the figure indicate the boundaries of the region for each of the 10 digits that contains 2/3 of the occurrences of the digit. These regions were located by calling the python function **dbscan( )** iteratively until the 2/3 criterion was satisfied. The boundary of the cluster was then drawn with the **ConvexHull( )** function, which additionally returns the volume of the enclosed region and enables the identification of points that fall inside the boundary. Advanced procedures such as dual classifier domain adversarial networks further separate the digit clusters and hence reduce the percentage of mislabeled data[102] while convolutional and related neural networks that are specifically adapted to image processing yield significantly higher prediction accuracies for a given computational overhead.[103]

Figure 2 displays for two different $x$-axis scales the accuracy of each digit, evaluated using the test dataset, as a function of epoch number as predicted by a linear neural network in which a 3 element input layer corresponding to the lowest-order 3 PCA components of the training data set is directly connected to a 10 element dense layer with a **softmax** activation function. The essential components of this computation, which are shared by the remaining calculations of this paper except where noted, are summarized by the pseudocode in the appendix. A batch size of 128 and the TensorFlow **RMSprop** optimizer with a learning rate of $1.5 \times 10^{-3}$ are employed (although a more efficient method could employ adaptive iterative learning-based procedures that continuously adjust the iterative learning gain such as in [6]) while the neural network parameters are randomly reinitialized at the beginning of each computation as in [96]. Further, graphs that display the accuracy of the digits as a function of epoch number are averaged over 30 separate calculations to reduce statistical fluctuations. In the left (2a) and right (2b) figures, the accuracy is evaluated from the test set at the end of each epoch and (in a separate calculation) after every 40 batches, respectively. Since each epoch comprises 469 batches, the digit accuracies in Figure 2(b) were calculated 11 times during each epoch. Note that employing the lowest-order PCA components as inputs in place of the actual $28 \times 28$ pixel image data greatly increases the accuracy differences among the digits and therefore the interpretability of the NN output. Network properties that are not evident from a standard calculation based on the full $28 \times 28$ MNIST pixel arrays can then be visualized, distinguishing our calculations from those of previous studies.

Figure 3 is generated by a nonlinear neural network with the 3-element input layer followed by dense 512 and 50 neuron layers, both employing **relu** activation functions and a 10 neuron **softmax** output layer. Figure 3(c) displays the equivalent result to Figure 3(b), but with a learning rate of $1.5 \times 10^{-4}$. The nonlinearity does not significantly influence the behavior of the accuracy evolution curves as the asymptotic behavior of the curves in Figure 2 and Figure 3 are qualitatively similar while the accuracy of the digits which, as noted below, is largely governed by the degree of overlap of the digit distributions in PCA space, is largely identical between the two figures. However, the calculations of Figure 3 require additional epochs compared to those of Figure 2 since the network contains numerous additional degrees of freedom that must be simultaneously optimized. The fluctuations in the accuracy evolution curves of Figures 3(b) and (c) are greater than those of the linear network, presumably because of the larger number of local minima in the nonlinear network loss function. Although the curves in Figure 3(c) are smoother because of the smaller learning rate, a larger number of training steps were employed.

Figure 4 and Figure 5 display results analogous to those of Figure 2 and Figure 3, respectively but for randomly rotated digits and the standard learning rate of $1.5 \times 10^{-3}$. Evidently only the digits 0 and 1 are efficiently resolved by the network in both the linear and nonlinear cases. Although the nonlinear network predictions are comparable to the nonrotated case, the digit accuracy converges more slowly, again presumably because of a greater number of local minimum in the loss function.

A salient feature of Figure 2 – Figure 5 is the accuracy of the digits 0 and 1 which are located along the lowest-order principal axis furthest from and on opposite sides of the origin in Figure 1. This is consistent with the lowest order principal axis coinciding with the direction of greatest variance while each PCA axis typically corresponds to a distinct, generally abstract geometric feature which is here the degree of circularity. The positioning of the least and most circular digits 1 and 0 at the edges of the combined digit distributions in the figure ensures a minimal overlap with the remaining digits as is also evident from the confusion matrix of Figure 6 for a linear neural network after 40 epochs in which the 3 lowest order PCA components each of the MNIST digits are employed as input. The off-diagonal confusion matrix elements for 0 and 1 are small, indicating that these digits are rarely misinterpreted.

Examining further the unrotated digit case of Figure 2 and Figure 3, after 0 and 1 the digits 3 and 7 converge most rapidly and exhibit the greatest accuracy. These latter digits are located at the outer limits of the second-lowest principal axis so that their digit distributions (clusters) are again effectively isolated from those of most of the remaining digits. Additionally, the projection of the input data for each digit on a given PCA axis is determined by the extent to which the geometrical feature associated with the axis is present in the digit. The digits that possess most or least of this feature exhibit greater accuracy and faster convergence with the number of epochs.

For the remaining digits, the predictive accuracy of the neural network is qualitatively dependent on the degree of overlap in PCA space between the distribution of a given digit with the distributions of geometrically similar digits. For example, in the similarity group consisting of 4, 7 and 9, the number 7 exhibits a far higher accuracy than 4 while 9 exhibits the smallest accuracy. In fact, Figure 7, which displays the normalized volume of the enclosed region in Figure 1 associated with the digit given by the row number that falls within the corresponding region of the digit specified by the column number, demonstrates that the clusters for the digits 7 and 9 and 4 and 9 overlap significantly in PCA space but do not overlap with other digits. This observation is also consistent with the confusion matrix of Figure 6, which qualitatively resembles Figure 7.

The source of the differing accuracies of the neural network predictions of 4, 7 and 9 is further clarified in Figure 8 which depicts the boundaries of the regions of large sample densities, calculated as in Figure 1, for the three digits 7 (orange), 9 (silver) and 4 (green) when all other digits are removed from the PCA input data. The vertical and horizontal axes in Figure 8(a) and (b) specify the second and third lowest-order and the lowest and second lowest order PCA components, respectively while Figure 8(c) contains a full three-dimensional plot of the digit distributions. Taken together, these diagrams confirm that the distribution of 9 overlaps significantly with those of both 4 and 7, indicating that the geometry of 9 exhibits features of both digits.

Employing exclusively the unrotated data records for the digits 4, 7 and 9 in a three-layer neural network calculation with 512, 50 neuron **relu** layers followed by a 3 neuron **softmax** output layer yields the test set digit accuracy curves of Figure 9 after averaging over 100 independent computations. While this result qualitatively resembles the evolution of these 3 digits in the full 10 digit calculation, the test accuracy is larger in Figure 9 for which the directions of the principal axes are exclusively determined by the properties of the three digits.

Since the distributions of the remaining digits, 2, 3, 5, 6 and 8 largely overlap, the off-diagonal confusion matrix elements for these digits are therefore large, decreasing the predictive accuracy of the network. Further, since the optimization procedure is less able to distinguish between these digits, their accuracy curves exhibit large fluctuations, especially during the first few epochs. This effect is more pronounced for the randomly rotated digits in Figure 4 and Figure 5. Here as later evident from Figure 12, only the numbers 0 and 1 do not significantly overlap with other digits while the PCA distributions of the remaining digits are nearly identical and the off-diagonal confusion matrix elements for these digits are therefore large. This results in slow convergence and significant accuracy fluctuations with batch number. The amplitude of these fluctuations considerably exceeds those of the non-rotated case again suggesting larger densities and magnitudes of local minima in the loss function.

The lowest accuracy nonrotated digits for the linear network, 5 and 8, are mapped close to the origin of PCA space and therefore overlap with numerous other digits as evident from the confusion matrix. Accordingly, the optimizer requires several additional epochs before these digits cease to be misinterpreted as a collection of the more easily distinguished, higher accuracy digits. For the nonlinear network, 5 and 9 are least accurate as their decision boundaries almost coincide.

**4: PCA based pattern discrimination and interpolation:** Each local region in the coordinate system of the two or three lowest-order principal components can be associated with a $28 \times 28$ pixel pattern in the physical space of the input data. To implement this quasi-inversion operation, the PCA volume of interest is divided into histogram bins and an average is taken over all the $28 \times 28$ pixel patterns with PCA components inside each bin. For unrotated digits, Figures 10(a) and (b) result when $20 \times 20$ two-dimensional histogram bins are employed that extend in Figure 10(a) from the smallest to the largest values of the data along the lowest (horizontal) and second lowest (vertical) order principal axes and similarly in Figure 10(b) for the second and third lowest-order axes. The intensity of these patterns proportional to the number of contributing data records. The patterns resulting from locations at which two or more digit regions overlap interpolate between the shapes of these digits, analogous to the behavior of variational autoencoders. [104–110] Another feature of variational autoencoders replicated in Figure 10 is that the low-order PCA axes can often be related to specific geometrical properties of the pixel patterns. For example, the slope of the digit 1 decreases in Figure 10(a) with the value of the second lowest-order component such that the second order principal component value for which the 1 is most vertical roughly coincides with the value for which the number 0 is circularly symmetric. Indeed, since the digit 1 is perfectly non-circular and hence can be mapped to e.g. a 2 component vector $(-a, 0)$ in two component PCA space while 0 is perfectly circular and thus is mapped to $(+b, 0)$, while none of the other geometrical properties of the digits is as distinct, these digits appear at the extremities of the digit distributions along the lowest-order PCA axis. Since no other digits overlap at these two extremities, the network predicts 1 and 0 with highest accuracy.

Figure 10(a) can also be compared with the corresponding variational autoencoder (VAE) result, Figure 2 of [108]. In both the VAE and PCA diagrams the 0 and 1 are on opposite sides of the horizontal axis while 7 is located at the outer (here bottom) edge of the VAE diagram and 3 appears near the top of the area below the 0 region, although it shares this region with 6 and 2. The numbers 5 and 8 are again situated in the middle of the VAE plot, indicating that they can be confused with other digits and with each other, reducing their accuracy in neural network calculations. As in the PCA case, 4, 7 and 9 are roughly adjacent.

To deconstruct this behavior further, in Figure 11 the PCA is applied only to the data records for the four digits furthest from the origin in the two-dimensional PCA space, namely 0, 1, 3 and 7. The two-dimensional histograms in these figures employ $32 \times 32$ bins between the largest and smallest values of the lowest and second lowest-order and of the second and third lowest-order components in Figures 11(a) and Figure 11(b) respectively. Note that in the regions bordering the 7 and 0 and the 0 and 3 regions in Figure 11(a) several overlapping patterns appear that somewhat resemble 9 and 5, while an approximation to 8 is present near the center of the distribution. This provides additional insight into the reduced accuracy of the digits 5, 8 and 9 relative to 0, 1, 3 and 7 in the neural network results of Figure 2. A broader implication, however, is that if, for example, input data records consisting of random perturbations of any four patterns that map to positions along the four positive and negative directions along the coordinate axes in two-dimensional PCA space generates a collection of artificial "digits" from the regions between these points. Further, if a set of patterns are identified that possess divergent geometric features, then once the lowest-order principal axes are determined from a collection of data records that are distortions of these patterns, projecting a new image onto these axes yields a point whose coordinates indicate the geometrical content of the image relative to the features of interest. This procedure could be employed more generally to quantify the degree of similarity between two images with differing complex geometries.

Regenerating Figure 10 but with randomly rotated input data yields Figure 12. While in Figure 10(a) and Figure 11(a) the second lowest-order component is related to the angle of inclination, the angle of inclination of the pixel patterns in Figure 12(b) instead varies with the polar angle. That is, the undistorted 0 digit appears at the PCA origin while the linear patterns corresponding to 1 that are most distorted by rotation are located along the outer perimeter of the distribution. Individual digits except for 1 and 0 are obscured in Figure 12 since many digits are superimposed within each histogram bin. This contrasts with the VAE for which, as noted in [108], the dominant output patterns are the circularly symmetric zero digit and a line resembling the digit 1 in various orientations, which results from the average over all digits at each angular displacement.

The PCA can also be employed to demonstrate the curvature of the decision boundaries associated with nonlinearity in a neural network. For example, in Figure 13 the second lowest-order PCA component is plotted against the lowest-order component for each element of the training data, where the colors indicate the digit predicted for each PCA point for (a) a linear neural network and (b) a nonlinear neural network consisting again of dense 512 and 50 **relu** layers followed by a standard 10 element **softmax** layer. Evidently, the nonlinear neural network decision boundaries describe curved lines while the boundaries are straight in the linear case, as also evident in e.g.[111] This elucidates the greater accuracy afforded by a nonlinear network. Further, if a PCA analysis is performed on the output of each nonlinear layer, the curvature presumably increases after each successive layer, which could be utilized as a metric for the evolution of nonlinearity through the network.

The probability distribution of the outputs in the final, **softmax** layer before the maximum likelihood discriminator is applied can be visualized by reducing the ten-dimensional space of these intermediate values to two or three PCA dimensions. The representative result of Figure 14 displays the distribution in the three-dimensional space formed from the first, second and third lowest order PCA components obtained from a linear neural network with the full $28 \times 28$ MNIST arrays as inputs. As expected, the digits with highest accuracy, namely 1, 0, 3 and 7 are again mapped to the regions furthest from the PCA space origin while the remaining digits occupy the central regions of the combined digit distribution. Note that a small fraction of the data falls on lines that join the four peak positions, although some segments such as the line from 0 to 1 are absent. This indicates that the neural network occasionally generates probabilities for two of the highest accuracy digits that sum to unity with certain exceptions such as 1 and 0 because of their fundamentally different geometries. Further calculations show that if the 3 lowest PCA components are employed as the input to the neural network in place of the $28 \times 28$ images, the data points fill a volume of PCA space between the highest accuracy peaks rather than being effectively confined to lines between pairs of peaks. This implies that a nonlinear network incorporates the influence of a larger number of similar digits in the decision process for a given data record than a linear network.

**5. PCA and data balancing:** To illustrate the practical utility of the concepts introduced in the above discussion a, to our knowledge novel, entropy expression and an associated data balancing procedure are formulated. The latter method is both simple to implement and automatically adjusts to the local degree of confusion in PCA space associated with the similarity of multiple classes.

In particular, for each bin, $m$, in the two-dimensional histogram of e.g. Figure 10(a) (or alternatively the corresponding multi-dimensional histogram in the space of the lowest PCA components) a "local PCA entropy" is defined as $E_m = \sum_{i=0}^{9} p_i^{(m)} \log p_i^{(m)}$, where $p_i^{(m)}$ is the probability of occurrence of the *i*:th digit in the $m$:th histogram bin. Note that this entropy differs fundamentally from global definitions based on the PCA dimensionality reduction of the complete data set [112,113], which are applicable to distributed effects such as phase transitions.

To balance a data set with the local PCA entropy the data records that map to each local PCA entropy bin can be duplicated or augmented a number of times given by a data balancing factor, $1 + \mathrm{floor}(n_{\mathrm{expansion}} E_m)$, where the integer $n_{\mathrm{expansion}}$ is here termed the "entropy expansion factor" and the function $\mathbf{floor}(\,)$ returns the largest integer smaller than its argument. Controlling the data balancing in this manner through the local PCA entropy ensures that the duplicated records are those that lead to the greatest confusion among the digit classes and therefore require the greatest attention by the network. The entropy expansion factor and the functional form of the data balancing factor are here determined empirically but could be learned in the context of a neural network calculation. After adding the duplicated records to the original data set, the data is shuffled. Finally, to compare this strategy to other procedures, 60,000 records are extracted to generate a training set with the same number of records as the standard MNIST training set, while the standard MNIST test set of 10,000 records is used for evaluation. Figure 15 – Figure 17 then display the evolution of the test accuracy of each digit for the 512, 50, 10 layer nonlinear neural network with a batch size of 256, a learning rate of $0.5 \times 10^{-3}$ and an interval of 100 batches between successive test accuracy evaluations. The RMSprop optimizer was employed with **clipnorm**=0.05 and **epsilon**=1.e-5, as these values were found to yield increased numerical stability. The statistical fluctuations were again reduced by averaging over 100 calculations.

Figures 15 displays the unbalanced, $n_{\mathrm{expansion}} = 0$ result while Figure 16 is the balanced result for $n_{expansion} = 2$, where a $10 \times 10$ histogram in two-dimensional PCA space extending from the minimum to the maximum values of the data along each axis is employed to determine the local entropy. Increasing the number of histogram bins to $32 \times 32$ yields curves that are effectively identical within the statistical error of the figures. In Figure 17, the three-dimensional PCA space was similarly divided into $10 \times 10 \times 10$ histogram bins and the balancing procedure applied with $n_{\mathrm{expansion}} = 2$. Increasing $n_{\mathrm{expansion}}$ beyond 2 does not significantly affect the results at least up to $n_{\mathrm{expansion}} = 6$.

Although the differences between the curves appear small on the scale of the graph, the increase in the level of precision and the convergence rate of the $n_{\mathrm{expansion}} = 2$ result relative to 100% is significant. Further, the distribution of the digit accuracies for $n_{\mathrm{expansion}} = 2$ is significantly narrower than that of the standard, unbalanced calculation. The three-dimensional result implies, at least in this problem, that any increase in accuracy resulting from the addition of the third dimension is offset by statistical errors arising from the smaller population of data points within the resulting histogram bins.

Figure 18 compares the error, defined as the difference between unity and the test accuracy, of the local PCA-based oversampling method with $10 \times 10 \times 10$ three-dimensional PCA histogram bins with the results of SMOTE and ADASYN for $k = 5$, where $k$ is the number of nearest neighbors employed to generate synthetic samples. Small variations in $k$, however, did not visibly affect the results. That the smallest errors are obtained with the local PCA entropy method indicates that the ability of the technique to identify the precise features that vary the most among the input records enables the mitigation of critical errors that would otherwise be difficult to address arising from data records associated with different classes but with similar values of these features.

**6. Discussion:** While the above analysis is confined to the simple benchmark example of the MNIST data set, several general features of our results should be noted. First, regarding computation time, note that a small number of eigenvectors with eigenvalues in the vicinity of a given value can be rapidly determined with iterative eigenvector solvers, even for large matrices.[114,115] However, while this step can be executed with minimal additional overhead, the performance of the local entropy metric will be highly problem-dependent since if multiple PCA components possess nearly equal eigenvalues, the PCA subspace dimension (e.g. the number of geometrical features) required to distinguish among different data samples can be large. This in turn leads to an unreasonably large number of histogram bins and hence local entropy evaluations. Such behavior resembles a phase transition, near which the number of significant PCA components significantly increases.[116] A possible exception to this behavior, however, would occur if the input classes differ from each other substantially with regard to the geometric features associated with the first 2 or 3 principal axes. For example, if the different classes are associated with images that are elongated along axes that are rotated by angles that are unique to each class, then if the one of the low-order principal axes is associated with the angle of rotation, the classes would be separated along this dimension even if the variances associated with other geometrical features are comparable.

Although as a result of the considerations above, the practicality of data balancing algorithms based on the local PCA entropy is expected to be highly problem-dependent, many other conceivable applications of such a local entropy metric can be envisioned. To give two possible implementations, as the extensive analysis in the first part of this paper indicates, the geometrical origins of the confusion among classes can be inferred from the locations of the regions of smallest and largest local PCA entropy in a space of several low-order PCA axes. This information could accordingly perhaps be employed to develop procedures for improving the fidelity of or for appropriately prefiltering the input data. Furthermore, the local PCA entropy could aid in quantifying the properties of intermediate layers in a neural network. For example, while the digit distribution of e.g. Figure 14 is quite involved, its local PCA entropy would presumably be far simpler since most regions containing samples are dominated by a single digit class. Therefore, the local PCA entropy could elucidate aspects of the neural network behavior more transparently than the sample distribution in PCA space itself. All such methods, however, will be intrinsically problem-dependent as their accuracy will again depend on the degree to which the data can be described by a small number of PCA components.

**7. Conclusions:** This paper has presented a detailed study of the properties of the MNIST data records in low-order PCA space. Based on these results, a local PCA entropy was defined to quantify the degree of uncertainty associated with different regions in low-order PCA space. To illustrate the utility of this concept, a simple and flexible procedure for balancing datasets was then introduced. While such techniques are largely restricted to systems that exhibit a small number of dominant (typically complex) geometric features, their algorithmic efficiency suggests that they could prove useful in certain application areas.

On a more fundamental level, several aspects of the low-order PCA representation of a dataset that could potentially be exploited to improve the performance of neural networks were analyzed. These include the ability to both distinguish between and interpolate among classes in the input data and to identify qualitatively salient geometric features of these classes. An analysis of the location and width of the distributions associated with the individual classes in the space of the lowest-order principal components was found largely to explain the variation in accuracy among these classes in neural network calculations as well as their convergence rates as a function of batch or epoch number. Furthermore, the shape of these distributions can be utilized to estimate the effects of nonlinearity in a neural network and this information could, at least in principle, be employed to optimize the neural network architecture.

**Acknowledgements:** The Natural Sciences and Engineering Research Council of Canada (NSERC) is acknowledged for financial support. [grant number RGPIN-03907-2020]

**Biography:** David Yevick (Ph.D. 1979, F. OSA, IEEE, APS) is a professor of physics at the University of Waterloo having been previously at Queen's University (Kingston), Penn State University, the University of Lund and the Institute of Optical Research, Stockholm. He has published over 200 articles in optical communications, physics and computational methods.

**Compliance with Ethical Standards:** The research leading to these results received funding from the Natural Sciences and Engineering Research Council of Canada (NSERC) under grant agreement number RGPIN-03907-2020. The authors have no relevant financial or non-financial interests to disclose.

**Data availability statement:** The data employed in this paper is freely available as the MNIST data set at, among other sites, https://github.com/cvdfoundation/mnist?search=1.

**Appendix:** The following pseudocode elucidates the procedure employed to evaluate the accuracies of the individual digits from the lowest-order PCA components of the data set, which is implemented within the framework of the program presented in section 2.5.1 of [96].

**read** MNIST data set from file
**normalize** data records
**evaluate** the **number_of_components** lowest-order PCA components of each data record
**for number_of_epochs** epochs
    **assign my_batch** from the lowest-order PCA components
    **insert** batch into current network model, e.g.
        **DenseLayer**
        **DenseLayer**
        **SoftmaxLayer**
    **evaluate** categorial cross entropy between predictions and labels
    **apply** backpropagation algorithm and adjust network parameters with **RMSprop** optimizer
    **every batch_interval** batches
        **apply** current network model to the full test data set
        **evaluate** categorial cross entropy between correct and predicted labels
        **store** digit accuracies

**References:**

[1] A. Voulodimos, N. Doulamis, A. Doulamis, E. Protopapadakis, Deep Learning for Computer Vision: A Brief Review, Comput Intell Neurosci 2018 (2018) 1–13. https://doi.org/10.1155/2018/7068349.

[2] K. Md. Hasib, M. Oli Ullah, M. Imran Nazir, A. Akter, M. Saifur Rahman, ICDP: An Improved Convolutional Neural Network Model to Detect Pneumonia from Chest X-Ray Images, Lecture Notes in Networks and Systems 867 LNNS (2024) 467–479. https://doi.org/10.1007/978-981-99-8937-9_32/TABLES/3.

[3] M.I. Nazir, A. Akter, M.A. Hussen Wadud, M.A. Uddin, Utilizing customized CNN for brain tumor prediction with explainable AI, Heliyon 10 (2024) e38997. https://doi.org/10.1016/J.HELIYON.2024.E38997.

[4] K.R. Chowdhary, Natural Language Processing, Fundamentals of Artificial Intelligence (2020) 603–649. https://doi.org/10.1007/978-81-322-3972-7_19.

[5] H. Tao, Y. Zheng, Y. Wang, J. Qiu, V. Stojanovic, Enhanced feature extraction YOLO industrial small object detection algorithm based on receptive-field attention and multi-scale features, Meas Sci Technol 35 (2024) 105023. https://doi.org/10.1088/1361-6501/AD633D.

[6] Z. Peng, X. Song, S. Song, V. Stojanovic, Spatiotemporal fault estimation for switched nonlinear reaction–diffusion systems via adaptive iterative learning, Int J Adapt Control Signal Process 38 (2024) 3473–3483. https://doi.org/10.1002/ACS.3885.

[7] X. Song, Z. Peng, S. Song, V. Stojanovic, Interval observer design for unobservable switched nonlinear partial differential equation systems and its application, International Journal of Robust and Nonlinear Control 34 (2024) 10990–11009. https://doi.org/10.1002/RNC.7553;WGROUP:STRING:PUBLICATION.

[8] A.C. I. Goodfellow, Y. Bengio, Deep Learning, MIT Press (2016). https://www.deeplearningbook.org/.

[9] G. Fang, X. He, Y. Weng, L. Fang, Texture Features Derived from Sentinel-2 Vegetation Indices for Estimating and Mapping Forest Growing Stock Volume, Remote Sens (Basel) 15 (2023). https://doi.org/10.3390/RS15112821.

[10] P. Kupidura, The comparison of different methods of texture analysis for their efficacy for land use classification in satellite imagery, Remote Sens (Basel) 11 (2019). https://doi.org/10.3390/RS11101233.

[11] R.M. Haralick, Statistical and structural approaches to texture, Proceedings of the IEEE 67 (1979) 786–804. https://doi.org/10.1109/PROC.1979.11328.

[12] C. Li, M. Wand, Combining Markov Random Fields and Convolutional Neural Networks for Image Synthesis, 2016 IEEE Conference on Computer Vision and Pattern Recognition (CVPR) (2016) 2479–2486. https://doi.org/DOI:10.48550/arXiv.1601.04589.

[13] J.B. Pelz, Morphological image segmentation by local granulometric size distributions, J Electron Imaging 1 (1992) 46. https://doi.org/10.1117/12.55174.

[14] S.G. Mallat, A Theory for Multiresolution Signal Decomposition: The Wavelet Representation, IEEE Trans Pattern Anal Mach Intell 11 (1989) 674–693. https://doi.org/10.1109/34.192463.

[15] T.S. Huang, G.J. Yang, G.Y. Tang, A Fast Two-Dimensional Median Filtering Algorithm, IEEE Trans Acoust 27 (1979) 13–18. https://doi.org/10.1109/TASSP.1979.1163188.

[16] B. Weisst, Fast median and bilateral filtering, ACM Trans Graph 25 (2006) 519–526. https://doi.org/10.1145/1141911.1141918.

[17] M. Kass, J. Solomon, Smoothed local histogram filters, ACM SIGGRAPH 2010 Papers, SIGGRAPH 2010 (2010). https://doi.org/10.1145/1778765.1778837.

[18] J. Cassetti, D. Delgadino, A. Rey, A.C. Frery, Entropy Estimators in SAR Image Classification, Entropy 24 (2022). https://doi.org/10.3390/E24040509.

[19] Y. Wei, L. Tao, Efficient histogram-based sliding window, Proceedings of the IEEE Computer Society Conference on Computer Vision and Pattern Recognition (2010) 3003–3010. https://doi.org/10.1109/CVPR.2010.5540049.

[20] M. Kim, Efficient histogram dictionary learning for text/image modeling and classification, Data Min Knowl Discov 31 (2017) 203–232. https://doi.org/10.1007/S10618-016-0461-2.

[21] A. Velichko, M. Belyaev, M.P. Wagner, A. Taravat, Entropy Approximation by Machine Learning Regression: Application for Irregularity Evaluation of Images in Remote Sensing, Remote Sens (Basel) 14 (2022). https://doi.org/10.3390/RS14235983.

[22] C.E. Shannon, A Mathematical Theory of Communication, Bell System Technical Journal 27 (1948) 379–423. https://doi.org/10.1002/J.1538-7305.1948.TB01338.X.

[23] A. Nicolaï, P. Delarue, P. Senet, Intrinsic Localized Modes in Proteins, Sci Rep 5 (2015) 1–11. https://doi.org/10.1038/SREP18128;SUBJMETA.

[24] P.M. Piaggi, M. Parrinello, Entropy based fingerprint for local crystalline order, Journal of Chemical Physics 147 (2017). https://doi.org/10.1063/1.4998408.

[25] F. García-Ramos, H. Li, Local entropy theory and applications, (2024). https://arxiv.org/pdf/2401.10012 (accessed September 20, 2025).

[26] E. Glasner, X. Ye, Local entropy theory, Ergodic Theory and Dynamical Systems 29 (2009) 321–356. https://doi.org/10.1017/S0143385708080309.

[27] H. Li, K. Liu, Local entropy theory, combinatorics, and local theory of Banach spaces, (2025). https://arxiv.org/pdf/2507.03338 (accessed September 20, 2025).

[28] A. Bavrina, V. Sergeyev, Fast Calculation of the Local Entropy of a Digital Image Using Machine Learning, Lecture Notes in Networks and Systems 1245 LNNS (2025) 186–195. https://doi.org/10.1007/978-3-031-81083-1_18/FIGURES/6.

[29] W.J. Lin, J.J. Chen, Class-imbalanced classifiers for high-dimensional data, Brief Bioinform 14 (2013) 13–26. https://doi.org/10.1093/BIB/BBS006.

[30] V. López, A. Fernández, S. García, V. Palade, F. Herrera, An insight into classification with imbalanced data: Empirical results and current trends on using data intrinsic characteristics, Inf Sci (N Y) 250 (2013) 113–141. https://doi.org/10.1016/J.INS.2013.07.007.

[31] B. Krawczyk, Learning from imbalanced data: open challenges and future directions, Progress in Artificial Intelligence 5 (2016) 221–232. https://doi.org/10.1007/S13748-016-0094-0.

[32] M. Galar, A. Fernández, E. Barrenechea, H. Bustince, F. Herrera, Ordering-based pruning for improving the performance of ensembles of classifiers in the framework of imbalanced datasets, Inf Sci (N Y) 354 (2016) 178–196. https://doi.org/10.1016/J.INS.2016.02.056.

[33] K. Borowska, M. Topczewska, New Data Level Approach for Imbalanced Data Classification Improvement, in: R. Burduk, K. Jackowski, M. Kurzynski, M. Wozniak, A. Zolnierek (Eds.), PROCEEDINGS OF THE 9TH INTERNATIONAL CONFERENCE ON COMPUTER RECOGNITION, CORES 2015, 2016: pp. 283–294. https://doi.org/10.1007/978-3-319-26227-7_27.

[34] G. Kovacs, An empirical comparison and evaluation of minority oversampling techniques on a large number of imbalanced datasets, Appl Soft Comput 83 (2019). https://doi.org/10.1016/j.asoc.2019.105662.

[35] S. Cabezuelo, R. González, D. Campo, R. Barbero, N. Mduma, Data Balancing Techniques for Predicting Student Dropout Using Machine Learning, Data 2023, Vol. 8, Page 49 8 (2023) 49. https://doi.org/10.3390/DATA8030049.

[36] G.E.A.P.A. Batista, R.C. Prati, M.C. Monard, A study of the behavior of several methods for balancing machine learning training data, ACM SIGKDD Explorations Newsletter 6 (2004) 20–29. https://doi.org/10.1145/1007730.1007735.

[37] M. Buda, A. Maki, M.A. Mazurowski, A systematic study of the class imbalance problem in convolutional neural networks, NEURAL NETWORKS 106 (2018) 249–259. https://doi.org/10.1016/j.neunet.2018.07.011.

[38] H. Ali, M.N. Mohd Salleh, R. Saedudin, K. Hussain, M.F. Mushtaq, Imbalance class problems in data mining: a review, Indonesian Journal of Electrical Engineering and Computer Science 14 (2019) 1552. https://doi.org/10.11591/ijeecs.v14.i3.pp1552-1563.

[39] H. He, E.A. Garcia, Learning from Imbalanced Data, IEEE Trans Knowl Data Eng 21 (2009) 1263–1284. https://doi.org/10.1109/TKDE.2008.239.

[40] T. Ryan Hoens, N. V. Chawla, Imbalanced datasets: From sampling to classifiers, Imbalanced Learning: Foundations, Algorithms, and Applications (2013) 43–59. https://doi.org/10.1002/9781118646106.CH3.

[41] J. Song, Y. Shen, Y. Jing, M. Song, Towards Deeper Insights into Deep Learning from Imbalanced Data, in: J. Yang, Q. Hu, M.M. Cheng, L. Wang, Q. Liu, X. Bai, D. Meng (Eds.), COMPUTER VISION, PT I, 2017: pp. 674–684. https://doi.org/10.1007/978-981-10-7299-4_56.

[42] L. Kemper, G. Vorhoff, B.U. Wigger, Predicting student dropout: A machine learning approach, European Journal of Higher Education 10 (2020) 28–47. https://doi.org/10.1080/21568235.2020.1718520.

[43] L. Aulck, N. Velagapudi, J. Blumenstock, J. West, Predicting Student Dropout in Higher Education, ArXiv (2017). https://doi.org/arXiv:1606.06364.

[44] T.M. Barros, P.A.S. Neto, I. Silva, L.A. Guedes, Predictive models for imbalanced data: A school dropout perspective, Educ Sci (Basel) 9 (2019). https://doi.org/10.3390/EDUCSCI9040275.

[45] Y. Xiao, J. Wu, Z. Lin, Cancer diagnosis using generative adversarial networks based on deep learning from imbalanced data, Comput Biol Med 135 (2021). https://doi.org/10.1016/j.compbiomed.2021.104540.

[46] S.F. Abdoh, M.A. Rizka, F.A. Maghraby, Cervical Cancer Diagnosis Using Random Forest Classifier With SMOTE and Feature Reduction Techniques, IEEE ACCESS 6 (2018) 59475–59485. https://doi.org/10.1109/ACCESS.2018.2874063.

[47] X. Yuan, L. Xie, M. Abouelenien, A regularized ensemble framework of deep learning for cancer detection from multi-class, imbalanced training data, Pattern Recognit 77 (2018) 160–172. https://doi.org/10.1016/j.patcog.2017.12.017.

[48] A. Karpatne, I. Ebert-Uphoff, S. Ravela, H.A. Babaie, V. Kumar, Machine Learning for the Geosciences: Challenges and Opportunities, IEEE Trans Knowl Data Eng 31 (2019) 1544–1554. https://doi.org/10.1109/TKDE.2018.2861006.

[49] A. Caté, L. Perozzi, E. Gloaguen, M. Blouin, Machine learning as a tool for geologists, Leading Edge 36 (2017) 215–219. https://doi.org/10.1190/TLE36030215.1.

[50] C.Y. Huang, H.L. Dai, Learning from class-imbalanced data: review of data driven methods and algorithm driven methods, DATA SCIENCE IN FINANCE AND ECONOMICS 1 (2021) 21–36. https://doi.org/10.3934/DSFE.2021002.

[51] L. Vu, D. Van Tra, Q.U. Nguyen, Learning from Imbalanced Data for Encrypted Traffic Identification Problem, in: PROCEEDINGS OF THE SEVENTH SYMPOSIUM ON INFORMATION AND COMMUNICATION (SOICT 2016), 2016: pp. 147–152. https://doi.org/10.1145/3011077.3011132.

[52] S. Wang, X. Yao, Using class imbalance learning for software defect prediction, IEEE Trans Reliab 62 (2013) 434–443. https://doi.org/10.1109/TR.2013.2259203.

[53] G. Cohen, M. Hilario, H. Sax, A. Hugonnet Stephane and Geissbuhler, Learning from imbalanced data in surveillance of nosocomial infection, Artif Intell Med 37 (2006) 7–18. https://doi.org/10.1016/j.artmed.2005.03.002.

[54] M. Saarela, O.-P. Ryynanen, S. Ayramo, Predicting hospital associated disability from imbalanced data using supervised learning, Artif Intell Med 95 (2019) 88–95. https://doi.org/10.1016/j.artmed.2018.09.004.

[55] P. Cichosz, S. Kozdrowski, S. Sujecki, Learning to Classify DWDM Optical Channels from Tiny and Imbalanced Data, ENTROPY 23 (2021) 1504. https://doi.org/10.3390/e23111504.

[56] Z. Rahman, A.M. Ami, A Transfer Learning Based Approach for Skin Lesion Classification from Imbalanced Data, in: PROCEEDINGS OF 2020 11TH INTERNATIONAL CONFERENCE ON ELECTRICAL AND COMPUTER ENGINEERING (ICECE), 2020: pp. 65–68. https://doi.org/10.1109/ICECE51571.2020.9393155.

[57] O.M. Olaitan, H.L. Viktor, SCUT-DS: Learning from Multi-class Imbalanced Canadian Weather Data, in: M. Ceci, N. Japkowicz, J. Liu, G.A. Papadopoulos, Z.W. Ras (Eds.), FOUNDATIONS OF INTELLIGENT SYSTEMS (ISMIS 2018), 2018: pp. 291–301. https://doi.org/10.1007/978-3-030-01851-1_28.

[58] X. Wan, J. Liu, W.K. Cheung, T. Tong, Learning to improve medical decision making from imbalanced data without a priori cost, BMC Med Inform Decis Mak 14 (2014). https://doi.org/10.1186/s12911-014-0111-9.

[59] Z. Li, J. Tang, F. Guo, Learning from real imbalanced data of 14-3-3 proteins binding specificity, Neurocomputing 217 (2016) 83–91. https://doi.org/10.1016/j.neucom.2016.03.093.

[60] X. Yang, Y. Zheng, M. Siddique, G. Beddoe, Learning from imbalanced data: a comparative study for Colon CAD - art. no. 69150R, in: M.L. Giger, N. Karssemeijer (Eds.), MEDICAL IMAGING 2008: COMPUTER-AIDED DIAGNOSIS, PTS 1 AND 2, 2008: p. R9150. https://doi.org/10.1117/12.770630.

[61] F. Deeba, S.K. Mohammed, F.M. Bui, K.A. Wahid, Learning from Imbalanced Data: A Comprehensive Comparison of Classifier Performance for Bleeding Detection in Endoscopic Video,

in: 2016 5TH INTERNATIONAL CONFERENCE ON INFORMATICS, ELECTRONICS AND VISION(ICIEV), 2016: pp. 1006–1009.

[62] J. Chen, R. Yang, C. Zhang, L. Zhang, Q. Zhang, DeepGly: A Deep Learning Framework With Recurrent and Convolutional Neural Networks to Identify Protein Glycation Sites From Imbalanced Data, IEEE ACCESS 7 (2019) 142368–142378. https://doi.org/10.1109/ACCESS.2019.2944411.

[63] F. Wotawa, H. Muehlburger, On the Effects of Data Sampling for Deep Learning on Highly Imbalanced Data from SCADA Power Grid Substation Networks for Intrusion Detection, in: 2021 IEEE 21ST INTERNATIONAL CONFERENCE ON SOFTWARE QUALITY, RELIABILITY AND SECURITY (QRS 2021), 2021: pp. 864–872. https://doi.org/10.1109/QRS54544.2021.00095.

[64] S.H. Khan, M. Hayat, M. Bennamoun, F.A. Sohel, R. Togneri, Cost-Sensitive Learning of Deep Feature Representations From Imbalanced Data, IEEE Trans Neural Netw Learn Syst 29 (2018) 3573–3587. https://doi.org/10.1109/TNNLS.2017.2732482.

[65] X. Liu, Y. Yao, Y. Ma, N. Yu, Privacy-preserving Cost-sensitive Federated Learning from Imbalanced Data, in: M. Reformat, D. Zhang, N. Bourbakis (Eds.), 2022 IEEE 34TH INTERNATIONAL CONFERENCE ON TOOLS WITH ARTIFICIAL, ICTAI, 2022: pp. 20–27. https://doi.org/10.1109/ICTAI56018.2022.00012.

[66] A. Braytee, W. Liu, P. Kennedy, A Cost-Sensitive Learning Strategy for Feature Extraction from Imbalanced Data, in: A. Hirose, S. Ozawa, K. Doya, K. Ikeda, M. Lee, D. Liu (Eds.), NEURAL INFORMATION PROCESSING, ICONIP 2016, PT III, 2016: pp. 78–86. https://doi.org/10.1007/978-3-319-46675-0_9.

[67] Y. Sun, M.S. Kamel, A.K.C. Wong, Y. Wang, Cost-sensitive boosting for classification of imbalanced data, Pattern Recognit 40 (2007) 3358. https://doi.org/10.1016/j.patcog.2007.04.009.

[68] Z.H. Zhou, X.Y. Liu, Training cost-sensitive neural networks with methods addressing the class imbalance problem, IEEE Trans Knowl Data Eng 18 (2006) 63. https://doi.org/10.1109/tkde.2006.17.

[69] E. At, M. Aljourf, F. Al-Mohanna, M. Shoukri, Classification of Imbalance Data using Tomek Link(T-Link) Combined with Random Under-sampling (RUS) as a Data Reduction Method, Global Journal of Technology and Optimization 01 (2016). https://doi.org/10.4172/2229-8711.S1111.

[70] Y. Dong, X. Wang, A New Over-Sampling Approach: Random-SMOTE for Learning from Imbalanced Data Sets, in: H. Xiong, W.B. Lee (Eds.), KNOWLEDGE SCIENCE, ENGINEERING AND MANAGEMENT, 2011: pp. 343–352.

[71] T. Wongvorachan, S. He, O. Bulut, A Comparison of Undersampling, Oversampling, and SMOTE Methods for Dealing with Imbalanced Classification in Educational Data Mining, Information (Switzerland) 14 (2023) 54. https://doi.org/10.3390/info14010054.

[72] P.S.M. Saladi, T. Dash, Genetic Algorithm-Based Oversampling Technique to Learn from Imbalanced Data, in: J.C. Bansal, K.N. Das, A. Nagar, K. Deep, A.K. Ojha (Eds.), SOFT COMPUTING

FOR PROBLEM SOLVING, SOCPROS 2017, VOL 1, 2019: pp. 387–397. https://doi.org/10.1007/978-981-13-1592-3_30.

[73] T.M. Barros, P.A.S. Neto, I. Silva, L.A. Guedes, Predictive Models for Imbalanced Data: A School Dropout Perspective, Education Sciences 2019, Vol. 9, Page 275 9 (2019) 275. https://doi.org/10.3390/EDUCSCI9040275.

[74] S.B. Kotsiantis, Supervised Machine Learning: A Review of Classification Techniques, Informatica 31 (2007) 249–268.

[75] M.O. Ullah, S.A. Raju, M.I. Nazir, A. Akter, M.S. Rahman, An Innovative Machine Learning Pipeline for Stroke Prediction on Imbalanced Data, 2023 International Conference on Information and Communication Technology for Sustainable Development, ICICT4SD 2023 - Proceedings (2023) 153–157. https://doi.org/10.1109/ICICT4SD59951.2023.10303319.

[76] B. Santoso, H. Wijayanto, K.A. Notodiputro, B. Sartono, Synthetic Over Sampling Methods for Handling Class Imbalanced Problems : A Review, IOP Conf Ser Earth Environ Sci 58 (2017) 012031. https://doi.org/10.1088/1755-1315/58/1/012031.

[77] N. V Chawla, K.W. Bowyer, L.O. Hall, W.P. Kegelmeyer, SMOTE: Synthetic minority over-sampling technique, JOURNAL OF ARTIFICIAL INTELLIGENCE RESEARCH 16 (2002) 321–357. https://doi.org/10.1613/jair.953.

[78] A. Fernandez, S. Garcia, F. Herrera, N. V Chawla, SMOTE for Learning from Imbalanced Data: Progress and Challenges, Marking the 15-year Anniversary, JOURNAL OF ARTIFICIAL INTELLIGENCE RESEARCH 61 (2018) 863–905. https://doi.org/10.1613/jair.1.11192.

[79] H. Han, W.Y. Wang, B.H. Mao, Borderline-SMOTE: A new over-sampling method in imbalanced data sets learning, in: D.S. Huang, X.P. Zhang, G.B. Huang (Eds.), ADVANCES IN INTELLIGENT COMPUTING, PT 1, PROCEEDINGS, 2005: pp. 878–887. https://doi.org/10.1007/11538059_91.

[80] M.I. Nazir, M. Oli Ullah, A. Akter, An Efficient Deep Transfer Learning based Apple Leaf Disease Classification, 2023 26th International Conference on Computer and Information Technology, ICCIT 2023 (2023). https://doi.org/10.1109/ICCIT60459.2023.10441087.

[81] G. Douzas, F. Bacao, Geometric SMOTE: Effective oversampling for imbalanced learning through a geometric extension of SMOTE, (2017). http://arxiv.org/abs/1709.07377.

[82] M. Lamari, N. Azizi, N.E. Hammami, A. Boukhamla, S. Cheriguene, N. Dendani, N.E. Benzebouchi, SMOTE–ENN-Based Data Sampling and Improved Dynamic Ensemble Selection for Imbalanced Medical Data Classification, Advances in Intelligent Systems and Computing 1188 (2021) 37–49. https://doi.org/10.1007/978-981-15-6048-4_4/TABLES/5.

[83] H. He, Y. Bai, E.A. Garcia, S. Li, ADASYN: Adaptive Synthetic Sampling Approach for Imbalanced Learning, in: 2008 IEEE INTERNATIONAL JOINT CONFERENCE ON NEURAL NETWORKS, VOLS 1-8, 2008: pp. 1322–1328. https://doi.org/10.1109/IJCNN.2008.4633969.

[84] J. Promrak, P. Kraipeerapun, S. Amornsamankul, Combining complementary neural network and error-correcting output codes for multiclass classification problems, in: 10th WSEAS International Conference on Applied Computer and Applied Computational Science, ACACOS'11, 2011: pp. 49–54. https://dl.acm.org/doi/10.5555/1965610.1965617.

[85] E.A. la Cal, J.R. Villar, P.M. Vergara, A. Herrero, J. Sedano, Design issues in Time Series dataset balancing algorithms, Neural Comput Appl 32 (2020) 1287–1304. https://doi.org/10.1007/s00521-019-04011-4.

[86] M.B.E. Silva, P.O. Broin, An optimised ensemble for antibody-mediated rejection status prediction in kidney transplant patients, in: 2020 IEEE Congress on Evolutionary Computation, CEC 2020 - Conference Proceedings, 2020. https://doi.org/10.1109/CEC48606.2020.9185739.

[87] S.K. Satapathy, S. Mishra, P.K. Mallick, G.S. Chae, ADASYN and ABC-optimized RBF convergence network for classification of electroencephalograph signal, Pers Ubiquitous Comput 27 (2023) 1161–1177. https://doi.org/10.1007/S00779-021-01533-4.

[88] S.A. Shilbayeh, S. Vadera, Cost-sensitive meta-learning framework, Journal of Modelling in Management 17 (2021) 987–1007. https://doi.org/10.1108/JM2-03-2021-0065.

[89] D. Thammasiri, D. Delen, P. Meesad, N. Kasap, A critical assessment of imbalanced class distribution problem, Expert Systems with Applications: An International Journal 41 (2014) 321–330. https://doi.org/10.1016/J.ESWA.2013.07.046.

[90] Q. Li, Y. Xie, A behavior-cluster based imbalanced classification method for credit card fraud detection, ACM International Conference Proceeding Series (2019) 134–139. https://doi.org/10.1145/3352411.3352433.

[91] L. Zhang, W.X. Wang, A Re-sampling method for class imbalance learning with credit data, in: Proceedings - 2011 International Conference of Information Technology, Computer Engineering and Management Sciences, ICM 2011, 2011: pp. 393–397. https://doi.org/10.1109/ICM.2011.34.

[92] E. Duman, A. Buyukkaya, I. Elikucuk, A novel and successful credit card fraud detection system implemented in a Turkish bank, Proceedings - IEEE 13th International Conference on Data Mining Workshops, ICDMW 2013 (2013) 162–171. https://doi.org/10.1109/ICDMW.2013.168.

[93] V. Shah, K. Passi, Data Balancing for Credit Card Fraud Detection Using Complementary Neural Networks and SMOTE Algorithm, (2021) 3–16. https://doi.org/10.1007/978-3-030-76776-1_1.

[94] P. Kraipeerapun, C.C. Fung, Binary classification using ensemble neural networks and interval neutrosophic sets, Neurocomputing 72 (2009) 2845–2856. https://doi.org/10.1016/J.NEUCOM.2008.07.017.

[95] P. Kraipeerapun, C.C. Fung, K.W. Wong, Uncertainty assessment using neural networks and interval neutrosophic sets for multiclass classification problems, WSEAS Transactions on Computers 6 (2007) 463–470. https://researchportal.murdoch.edu.au/esploro/outputs/journalArticle/Uncertainty-assessment-using-neural-networks-and/991005541072807891.

[96] F. Chollet, Deep Learning with Python, Second Edition, 2nd ed, Manning, Shelter Island, 2021.

[97] T.H. Chan, K. Jia, S. Gao, J. Lu, Z. Zeng, Y. Ma, PCANet: A Simple Deep Learning Baseline for Image Classification?, IEEE Transactions on Image Processing 24 (2015). https://doi.org/10.1109/TIP.2015.2475625.

[98] W. Khan, M. Turab, W. Ahmad, S.H. Ahmad, K. Kumar, B. Luo, Data Dimension Reduction makes ML Algorithms efficient, in: ICETECC 2022 - International Conference on Emerging Technologies in Electronics, Computing and Communication, 2022. https://doi.org/10.1109/ICETECC56662.2022.10069527.

[99] R. Sheikh, M. Patel, A. Sinhal, Recognizing MNIST Handwritten Data Set Using PCA and LDA, in: International Conference on Artificial Intelligence: Advances and Applications 2019, 2020: pp. 169–177. https://doi.org/10.1007/978-981-15-1059-5_20.

[100] L. Milosheski, G. Cerar, B. Bertalanic, C. Fortuna, M. Mohorcic, Deep Feature Learning for Wireless Spectrum Data, in: 2023 IEEE International Mediterranean Conference on Communications and Networking, MeditCom 2023, 2023: pp. 121–126. https://doi.org/10.1109/MeditCom58224.2023.10266604.

[101] M. Al-Hamadani, Classification and analysis of the MNIST dataset using PCA and SVM algorithms, Vojnotehnicki Glasnik 71 (2023) 221–238. https://doi.org/10.5937/VOJTEHG71-42689.

[102] Y. Sun, H. Tao, V. Stojanovic, Pseudo-label guided dual classifier domain adversarial network for unsupervised cross-domain fault diagnosis with small samples, Advanced Engineering Informatics 64 (2025) 102986. https://doi.org/10.1016/J.AEI.2024.102986.

[103] A. Azgar, M. Imran Nazir, A. Akter, M. Saddam Hossain, M. Anwar Hussen Wadud, M. Reazul Islam, MNIST Handwritten Digit Recognition Using a Deep Learning-Based Modified Dual Input Convolutional Neural Network (DICNN) Model, Lecture Notes in Networks and Systems 1014 LNNS (2024) 563–573. https://doi.org/10.1007/978-981-97-3562-4_44/TABLES/1.

[104] D.P. Kingma, M. Welling, An introduction to variational autoencoders, Foundations and Trends in Machine Learning 12 (2019) 307–392. https://doi.org/10.1561/2200000056.

[105] D.P. Kingma, M. Welling, Auto-encoding variational Bayes, in: 2nd International Conference on Learning Representations, ICLR 2014 - Conference Track Proceedings, International Conference on Learning Representations, ICLR, 2014. https://doi.org/10.61603/ceas.v2i1.33.

[106] M. Rolínek, D. Zietlow, G. Martius, M. Rolinek, D. Zietlow, G. Martius, Variational autoencoders pursue PCA directions (by accident), Proceedings of the IEEE Computer Society Conference on Computer Vision and Pattern Recognition 2019-June (2019) 12398–12407. https://doi.org/10.1109/CVPR.2019.01269.

[107] D. Yevick, Variational autoencoder analysis of Ising model statistical distributions and phase transitions, European Physical Journal B 95 (2022). https://doi.org/10.1140/epjb/s10051-022-00296-y.

[108] D. Yevick, Rotated Digit Recognition by Variational Autoencoders with Fixed Output Distributions, (2022). https://arxiv.org/abs/2206.13388v1.

[109] C. Doersch, Tutorial on Variational Autoencoders, (2021). http://arxiv.org/abs/1606.05908.

[110] Q. Fournier, D. Aloise, Empirical comparison between autoencoders and traditional dimensionality reduction methods, in: Proceedings - IEEE 2nd International Conference on Artificial Intelligence and Knowledge Engineering, AIKE 2019, 2019. https://doi.org/10.1109/AIKE.2019.00044.

[111] Y. Park, C.D. Kim, G. Kim, Variational Laplace autoencoders, in: 36th International Conference on Machine Learning, ICML 2019, 2019: pp. 8844–8856. arxiv: 2211.17267.

[112] R.K. Panda, R. Verdel, A. Rodriguez, H. Sun, G. Bianconi, M. Dalmonte, Non-parametric learning critical behavior in Ising partition functions: PCA entropy and intrinsic dimension, SciPost Physics Core 6 (2023). https://doi.org/10.21468/SciPostPhysCore.6.4.086.

[113] K. Greenewald, B. Kingsbury, Y. Yu, High-Dimensional Smoothed Entropy Estimation via Dimensionality Reduction, in: IEEE International Symposium on Information Theory - Proceedings, 2023. https://doi.org/10.1109/ISIT54713.2023.10206641.

[114] David. Yevick, A first course in computational physics and object-oriented programming with C++, Cambridge University Press, 2005.

[115] D. Yevick, A Short Course in Computational Science and Engineering, 2012. https://doi.org/10.1017/cbo9781139022262.

[116] D. Yevick, Conservation laws and spin system modeling through principal component analysis, Comput Phys Commun 262 (2021). https://doi.org/10.1016/j.cpc.2021.107832.

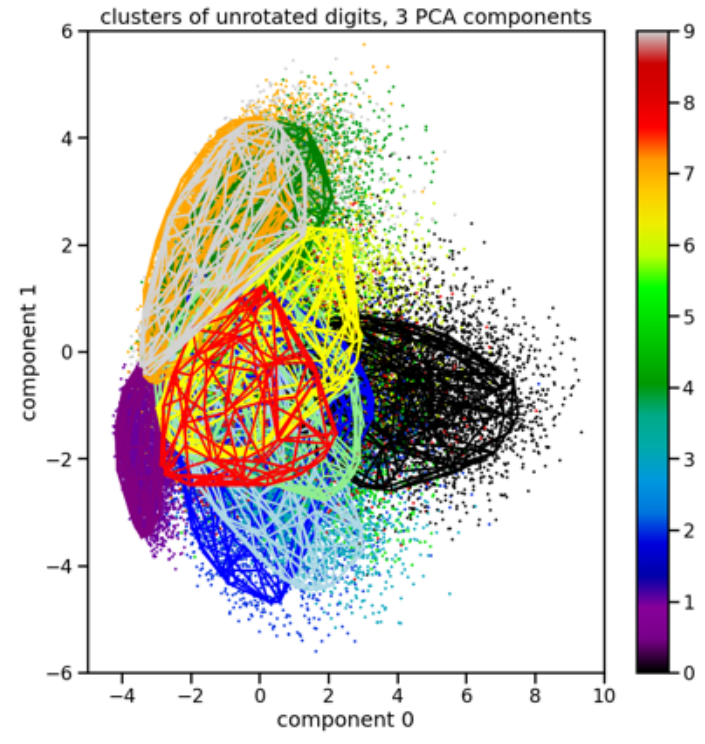

*Figure 1*: The second-order plotted against the first, lowest-order PCA component for the unrotated MNIST digits (dots) together with the boundaries of 2/3 of the data for each digit (solid lines).

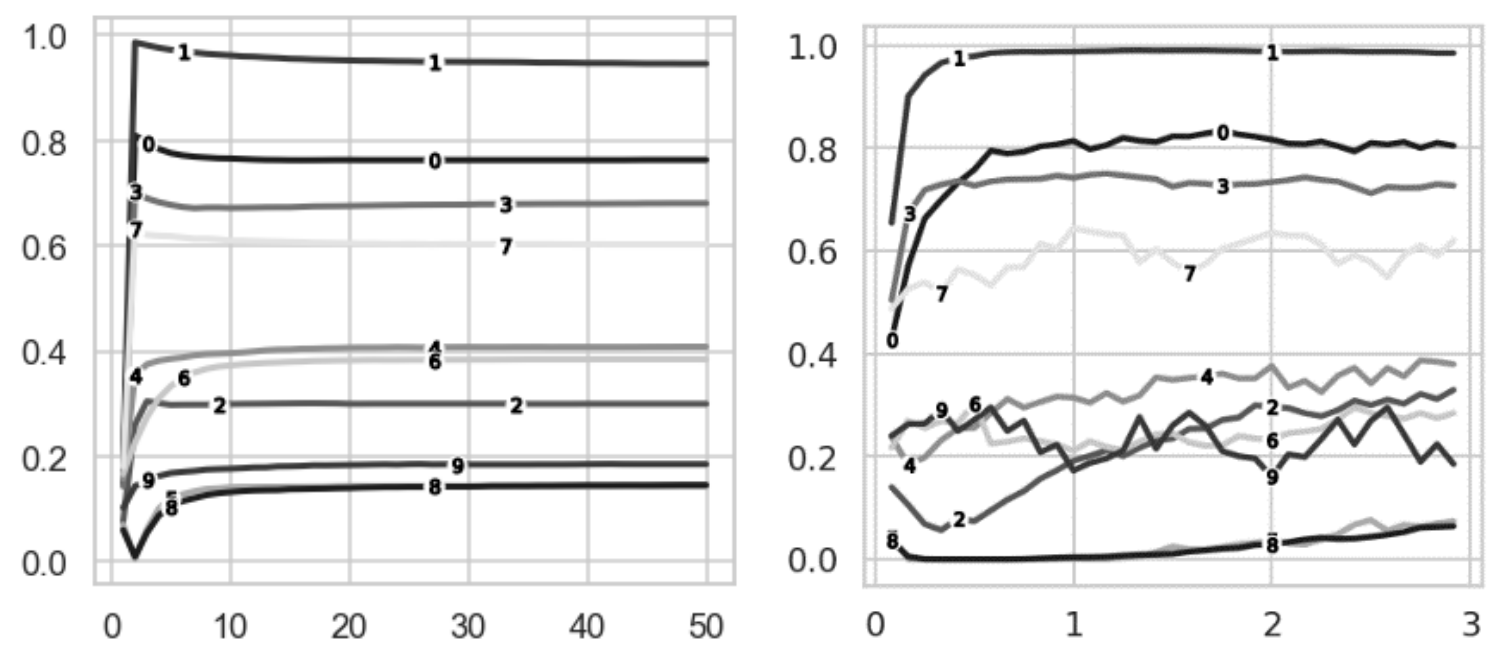

*Figure 2*: The prediction accuracy of each digit as a function of epoch number for a linear network where the input records are the 3 lowest order PCA components of each MNIST digit.

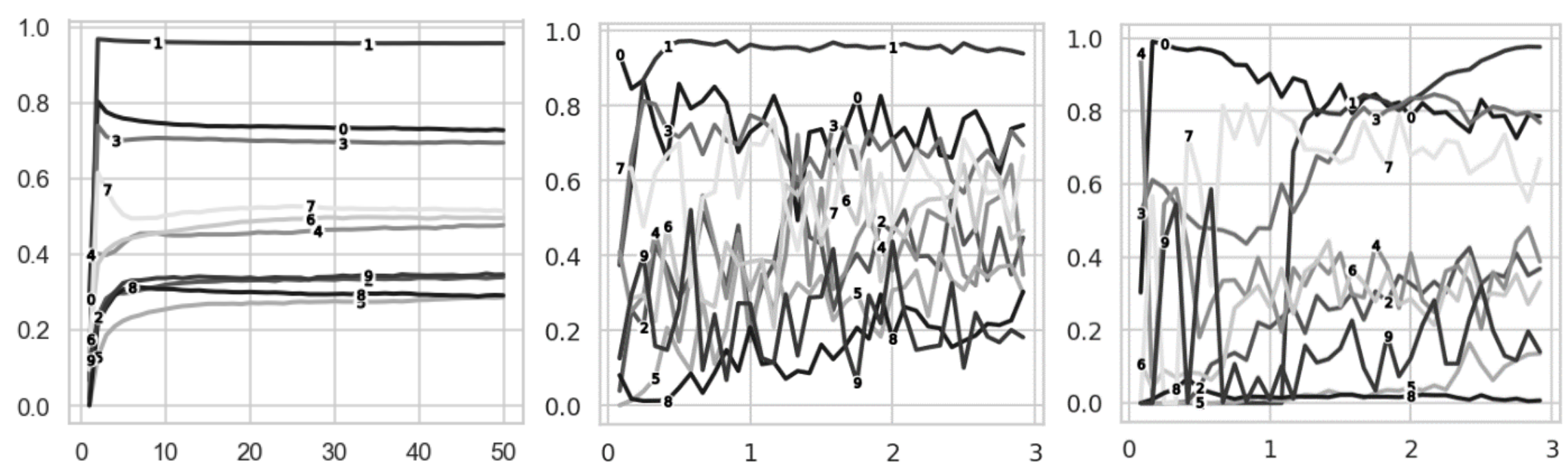

*Figure 3*: As in *Figure 2*(a) and (b) but for a dense nonlinear neural network with 512, 50 and 10 neuron layers. In Figure (c) the learning rate equals $1.5 \times 10^{-4}$.

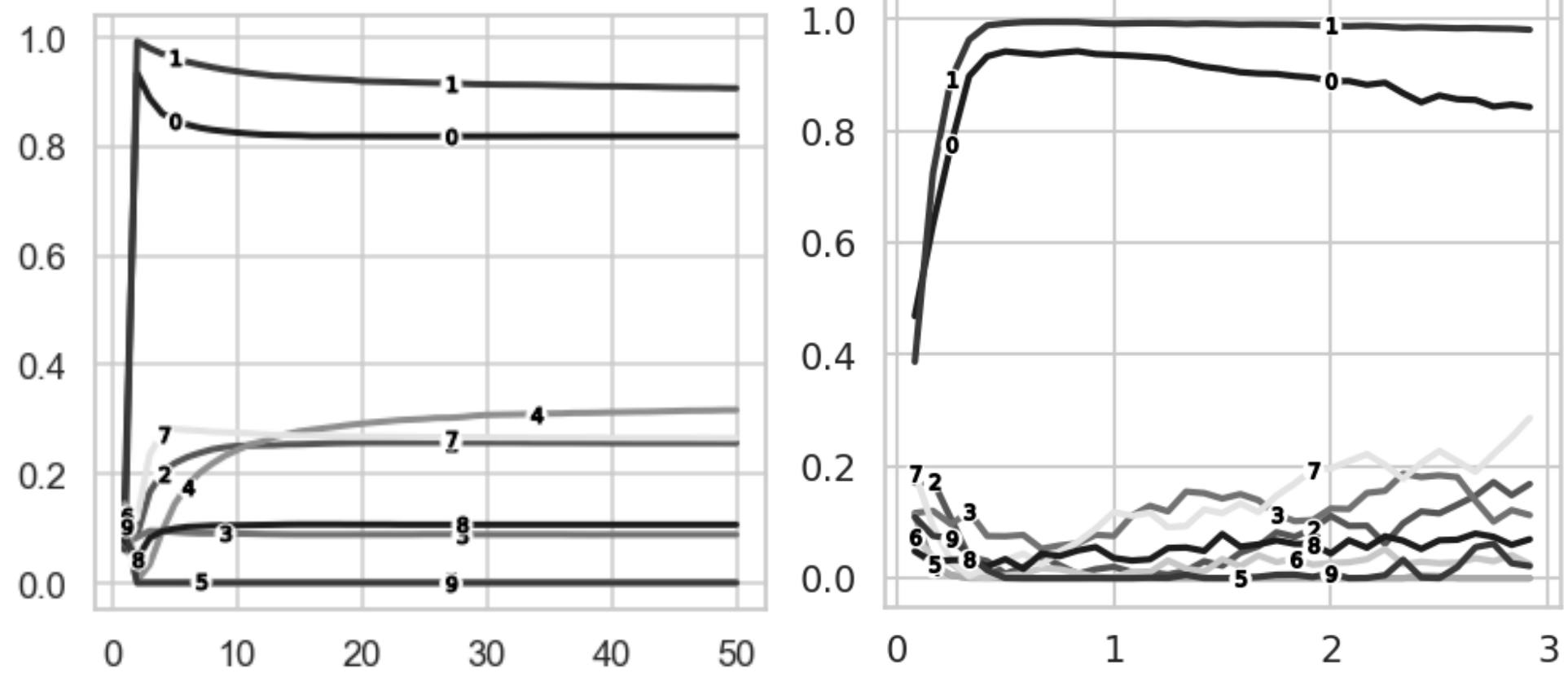

*Figure 4*: As in *Figure 2* but for randomly rotated digits.

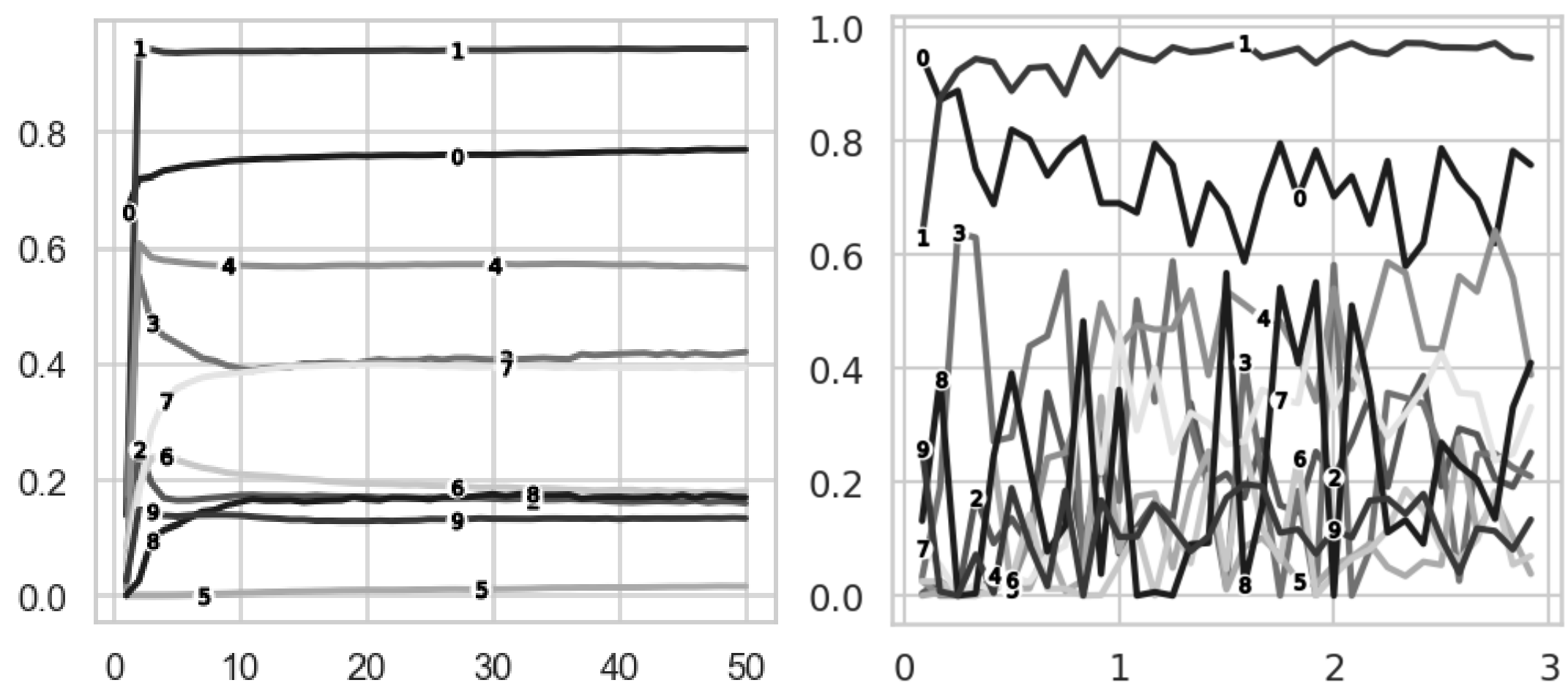

*Figure 5*: As in *Figure 3*(a) and (b) but for randomly rotated digits.

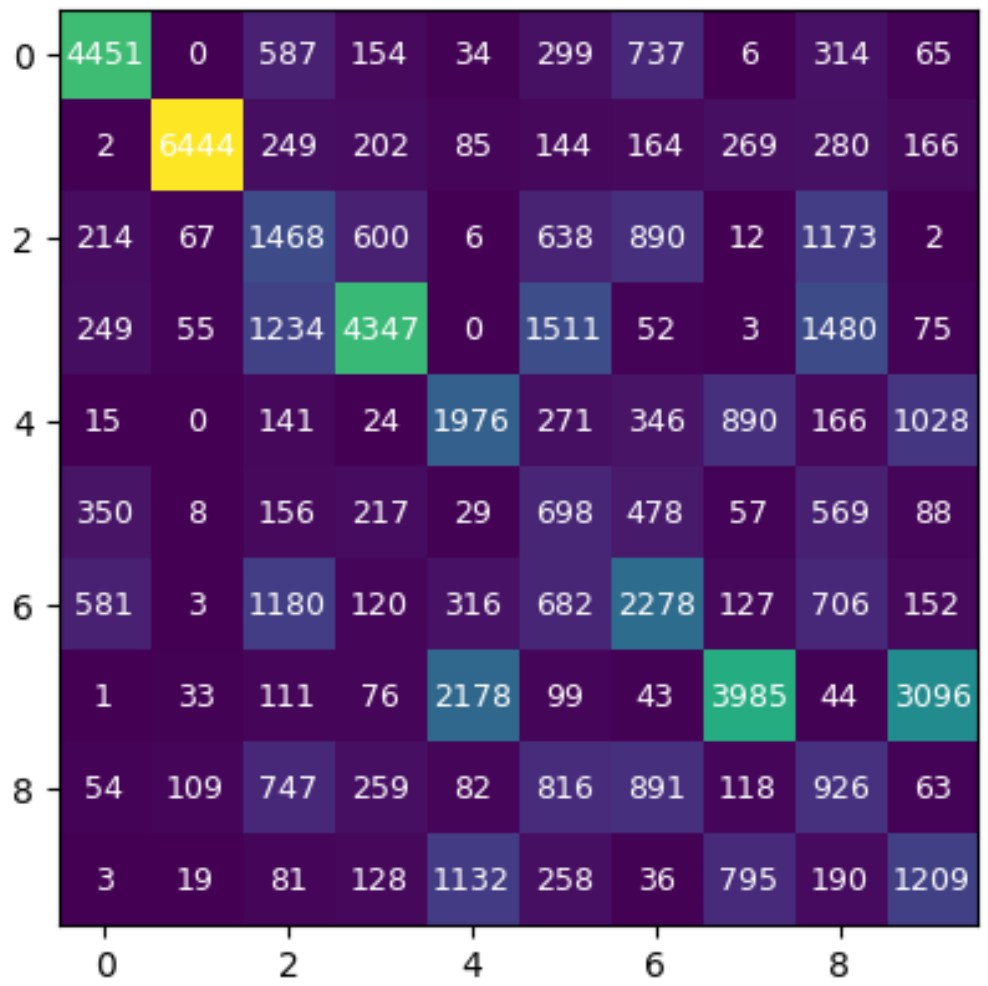

*Figure 6*: The confusion matrix after 40 epochs for the linear network of *Figure 2* with the lowest 3 MNIST PCA components as input

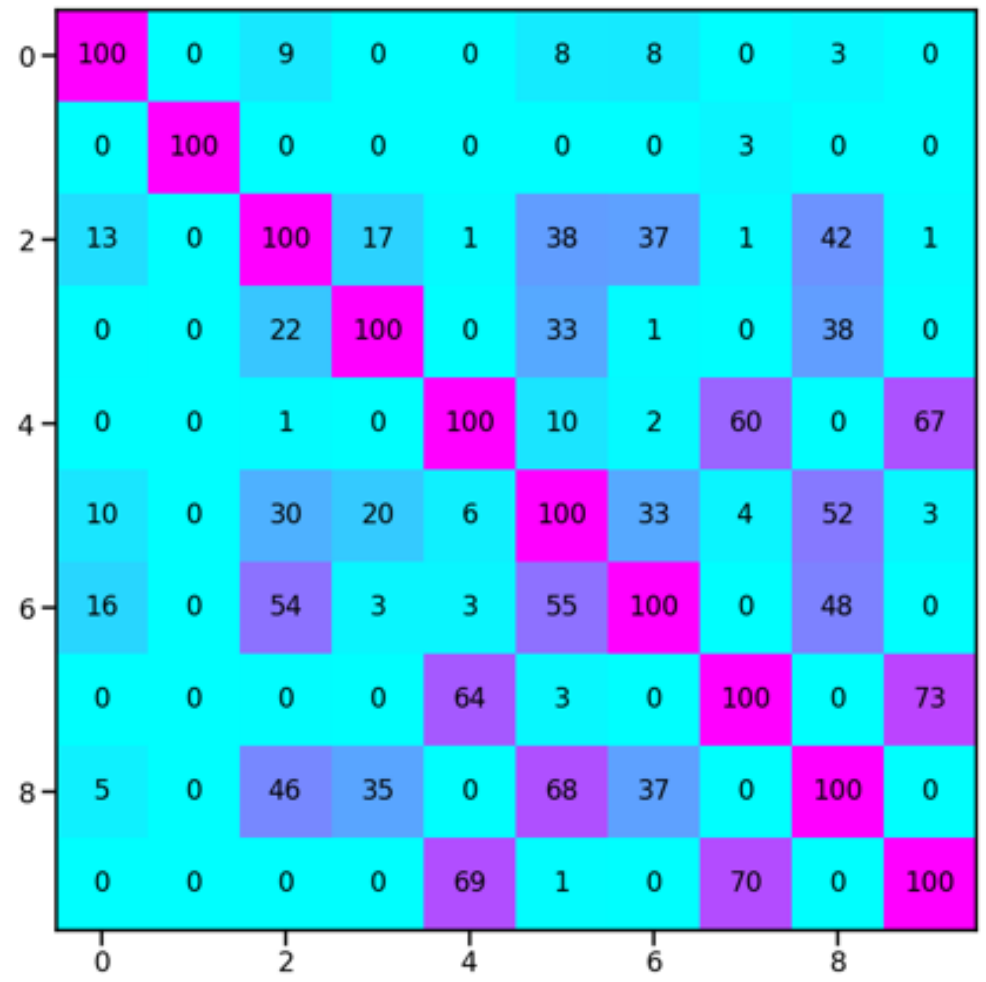

*Figure 7*: The normalized volume of the enclosed region in Figure 1 associated with the row number digit included within the column number digit region.

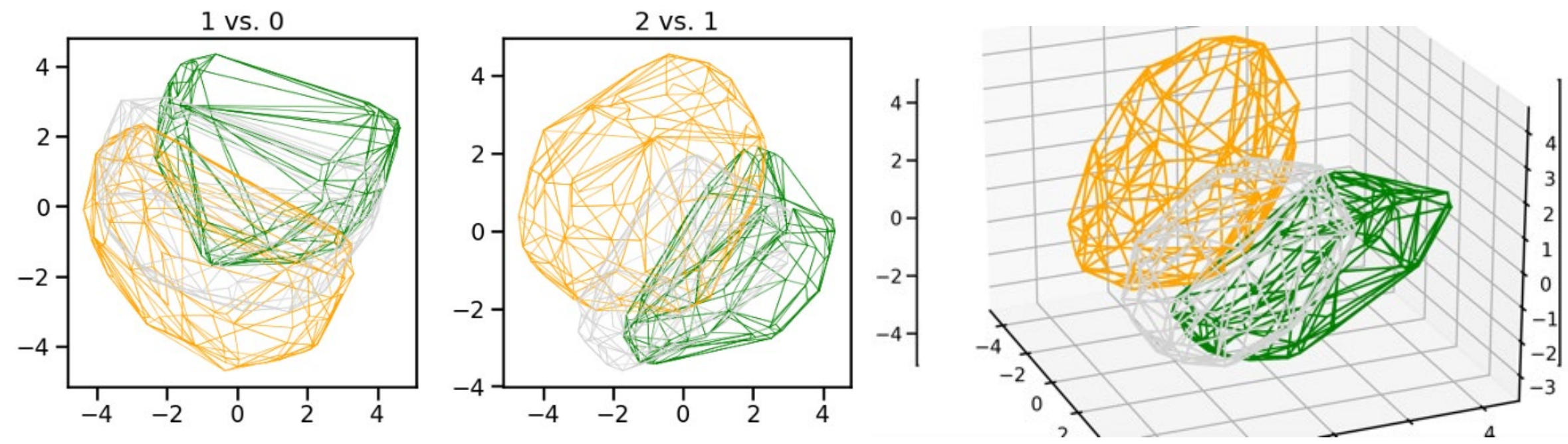

*Figure 8*: The cluster boundaries for 7 (orange), 9 (silver) and 4 (green) when all other digits are absent from the input. The vertical and horizontal axes in *Figure 8* (a) and (b) are the second and first lowest-order and the third and second lowest-order PCA components, respectively. *Figure 8*(c) is a three-dimensional plot.

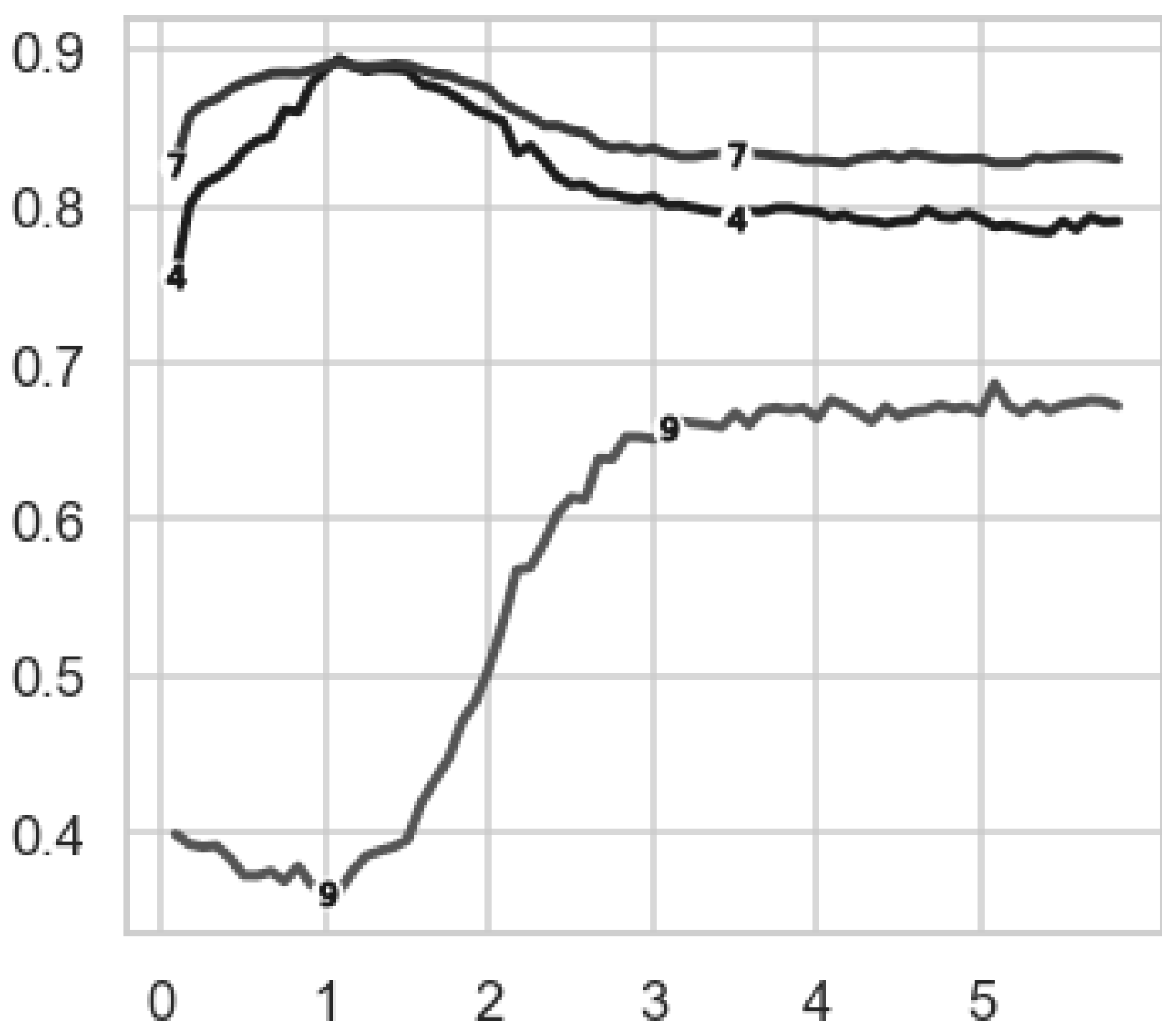

*Figure 9*: The digit accuracies for 4, 7 and 9 in a 3 layer nonlinear neural network calculation when other digits are absent.

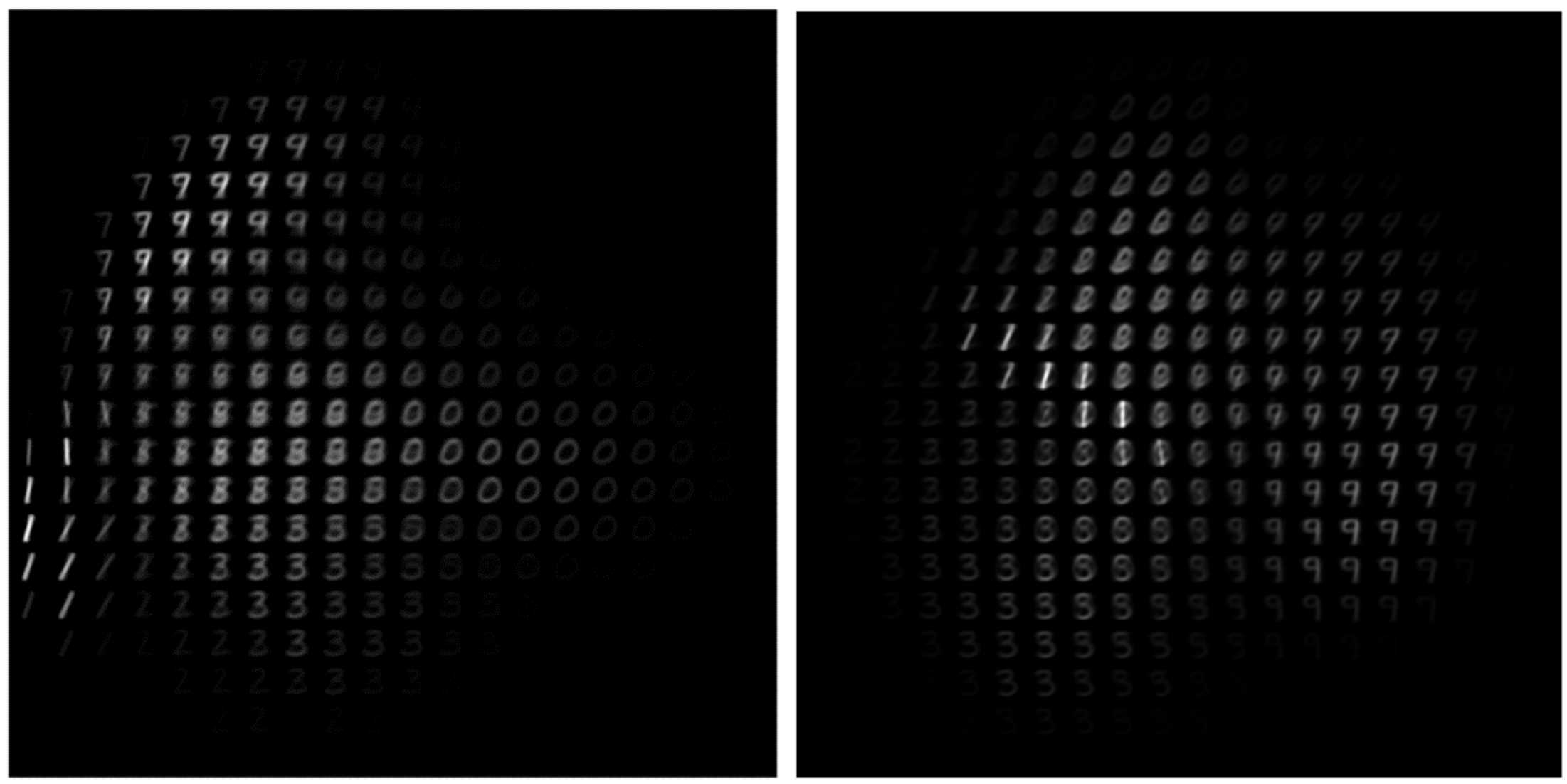

*Figure 10*: The inversion of the PCA space mapping for unrotated digits, and $20 \times 20$ two-dimensional histogram bins.

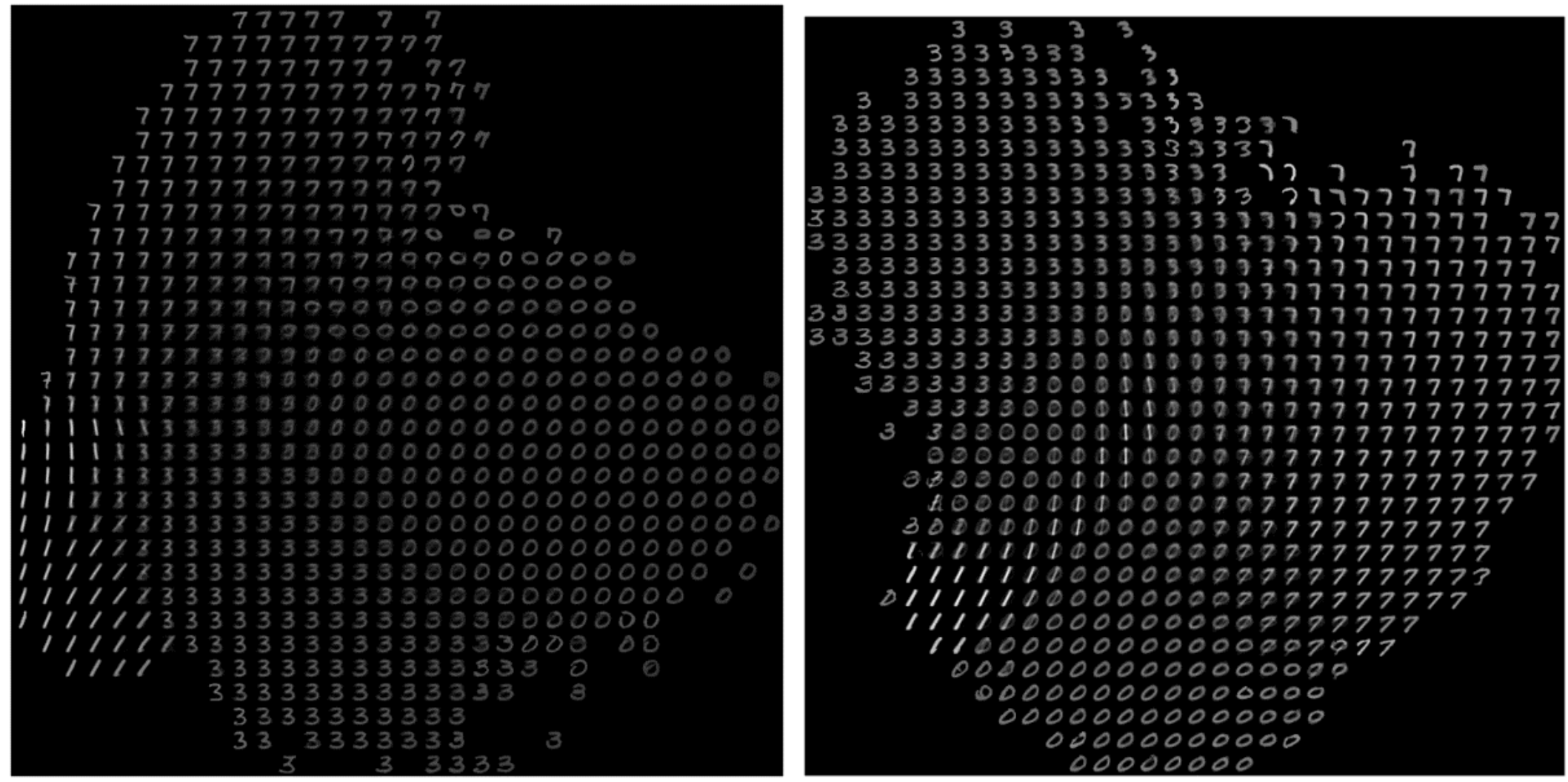

*Figure 11*: As in *Figure 10*, but when only 0, 1, 3 and 7 are present and for $20 \times 20$ uniformly normalized histogram bins.

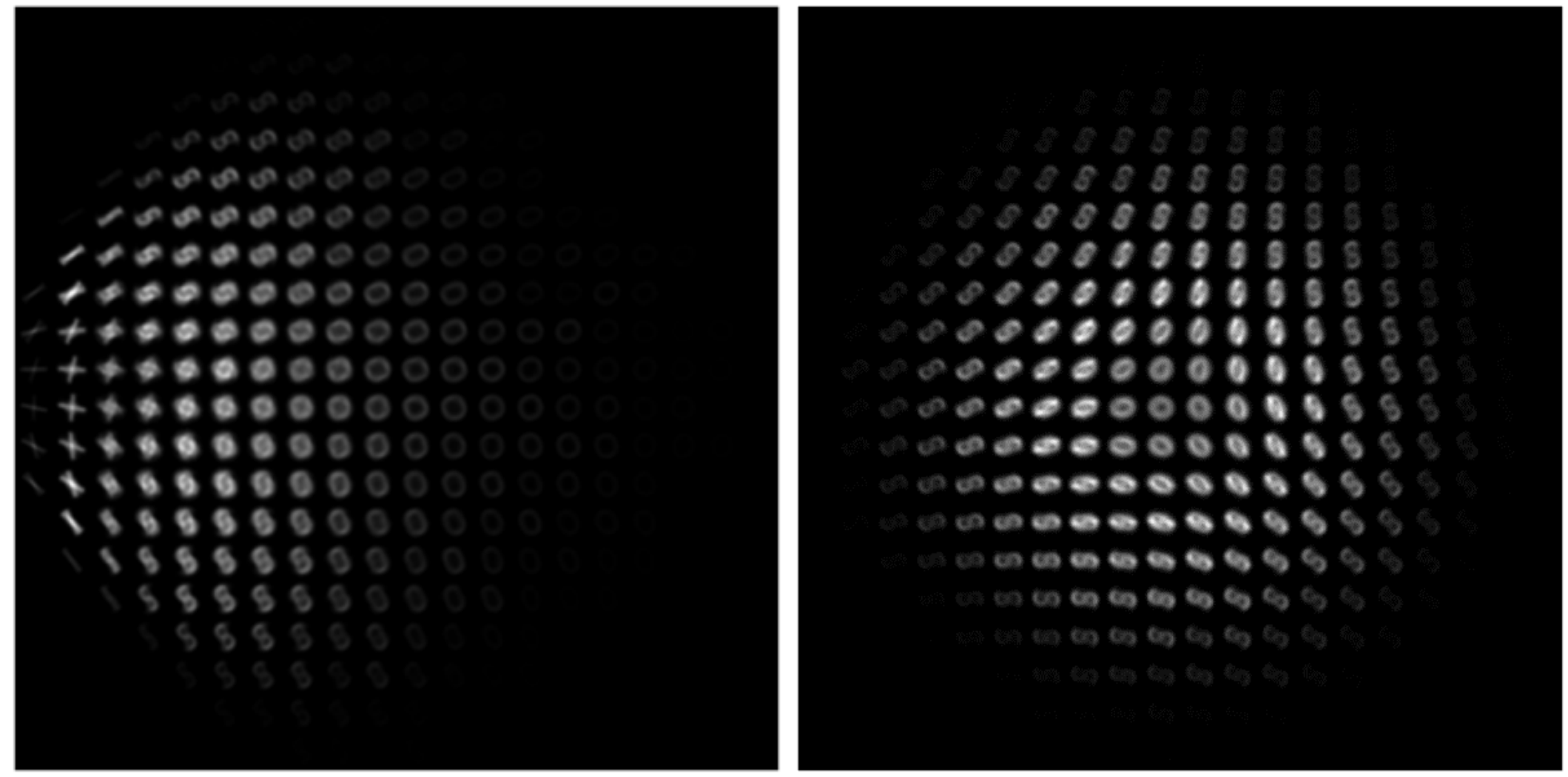

*Figure 12*: As in *Figure 10*, but for randomly rotated digits

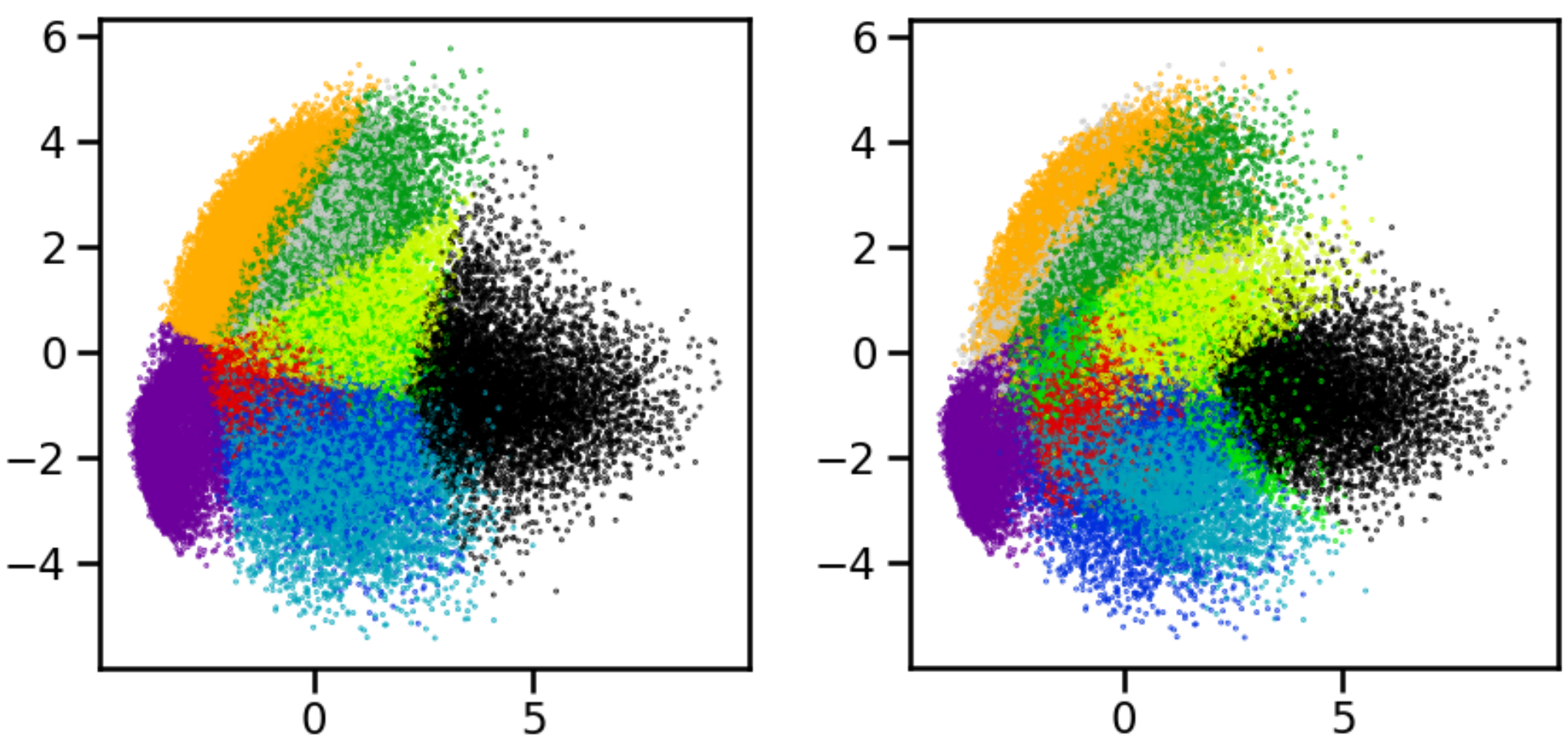

*Figure 13*: The second lowest-order PCA component plotted against the lowest-order component for the training data for a (a) linear neural network and (b) nonlinear neural network

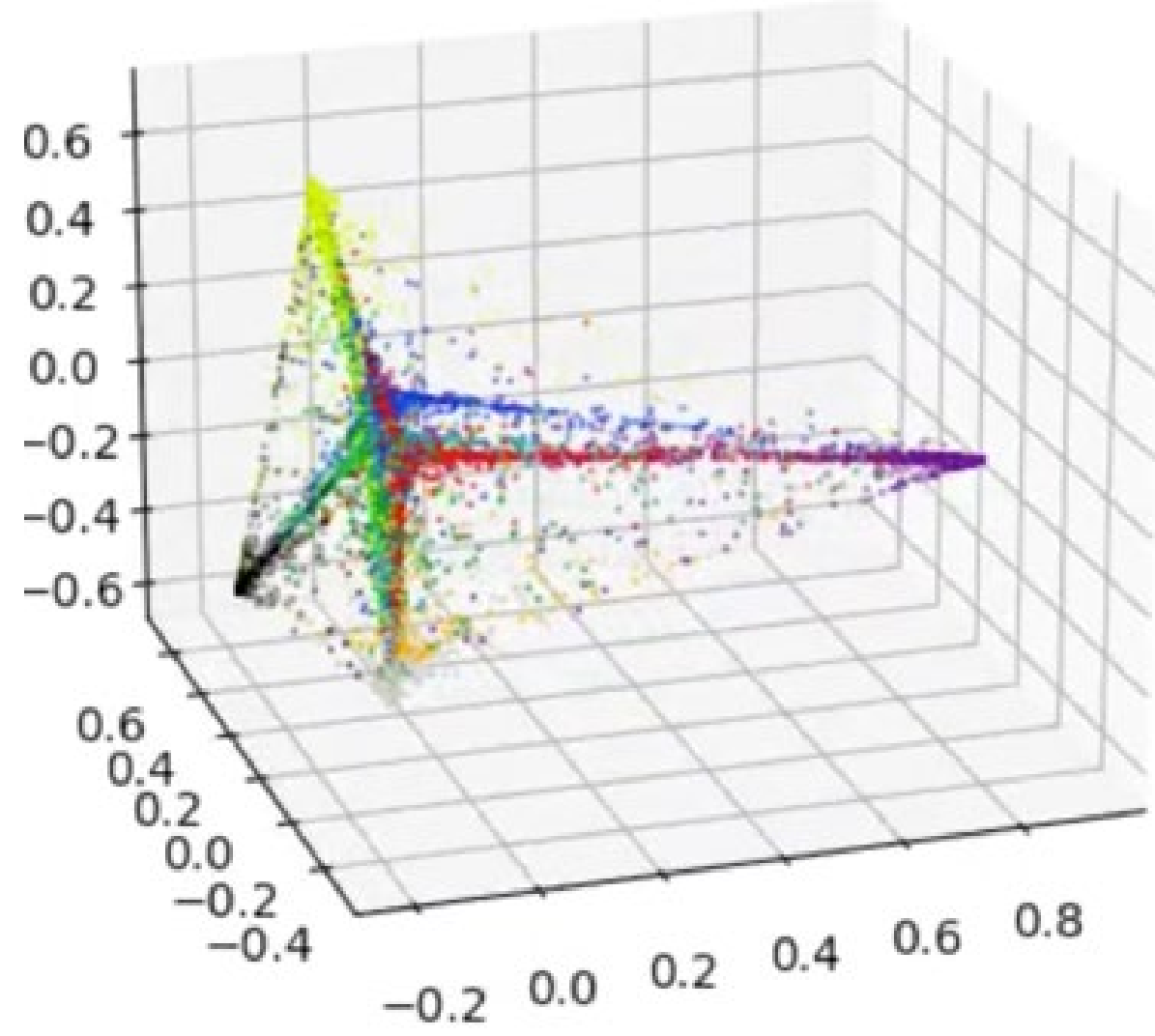

*Figure 14*: The distribution in PCA space in the last network layer before the linear discriminator for a linear network and the $28 \times 28$ pixel NMIST input data as input.

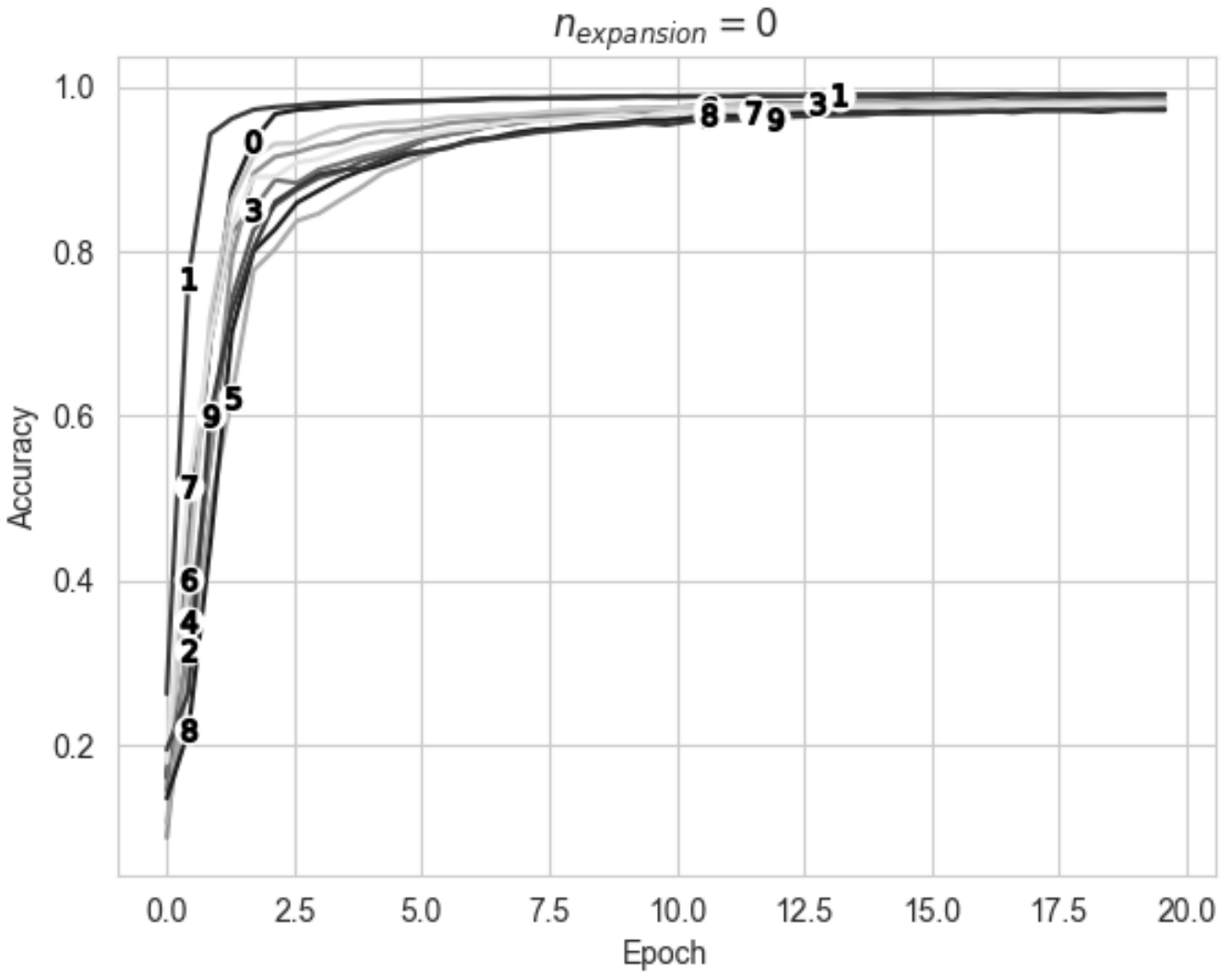

*Figure 15*: The accuracy of a standard nonlinear neural network calculation for the MNIST data set (equivalent to $n_{expansion} = 0$)

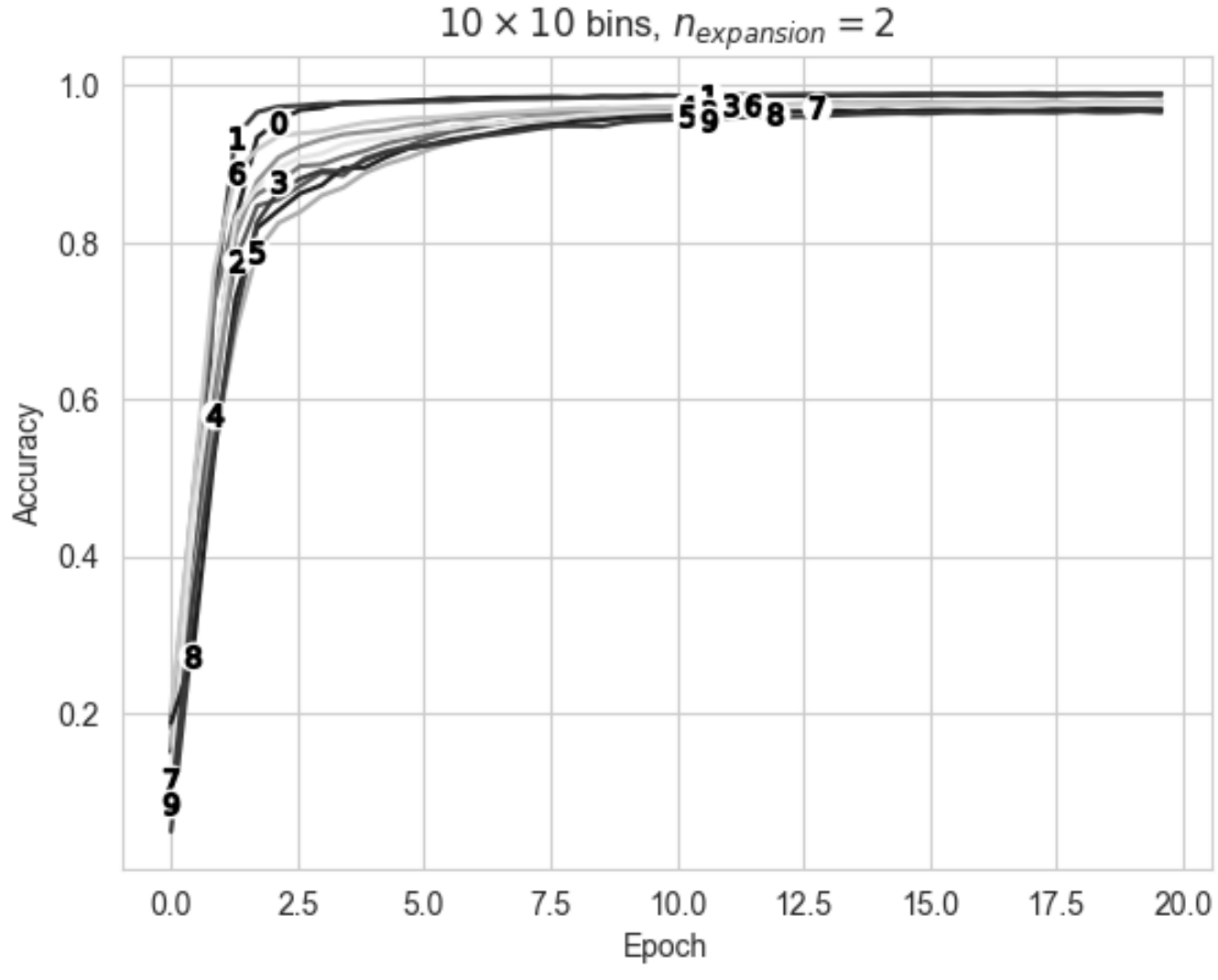

*Figure 16*: Same as *Figure 15* but with $n_{expansion} = 2$ , where the space spanned by the lowest 2 PCA axes is divided into a $10 \times 10$ histogram.

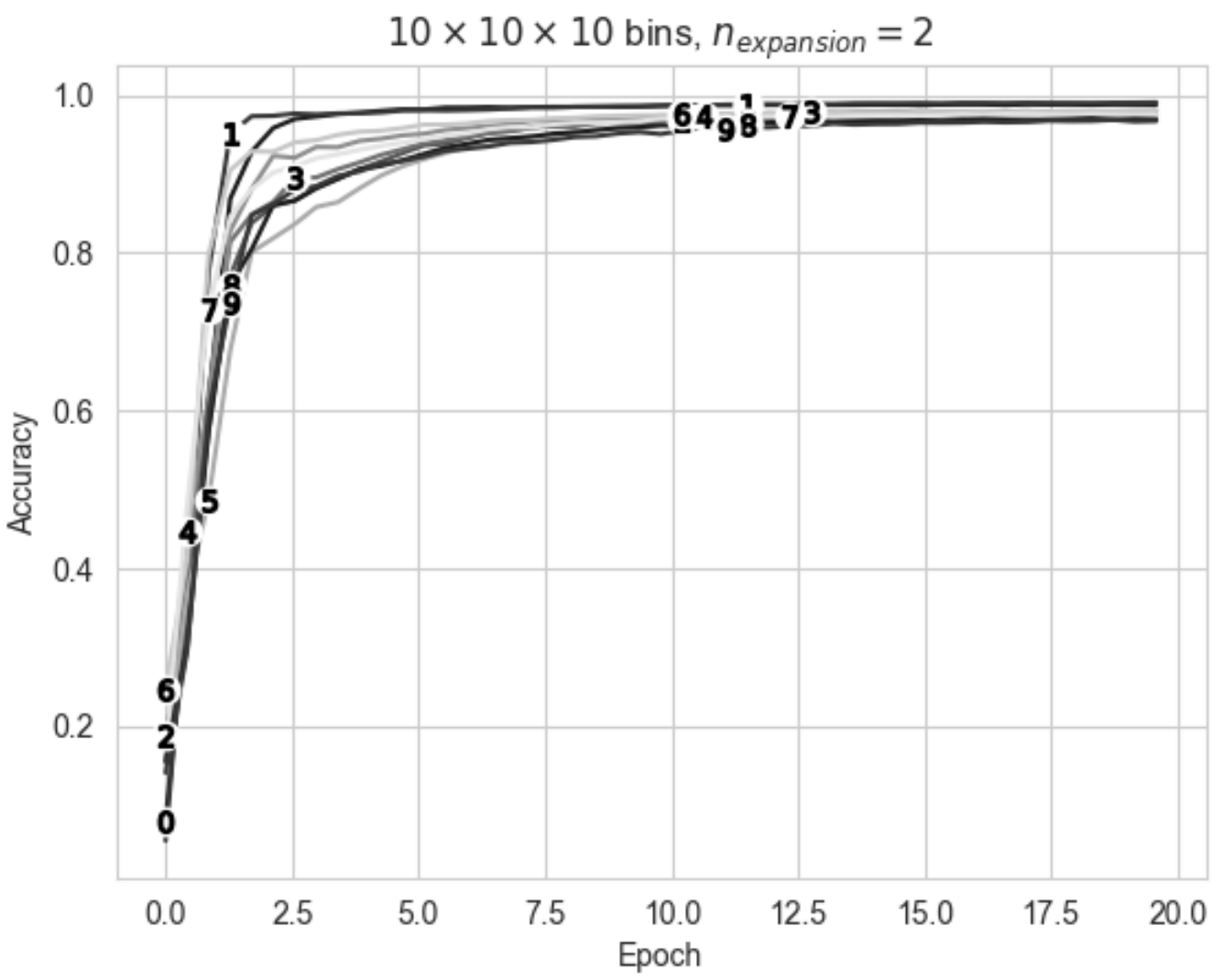

*Figure 17*: The $n_{expansion} = 2$ result of *Figure 16* but instead employing the lowest 3 PCA axes and a $10 \times 10 \times 10$ histogram.

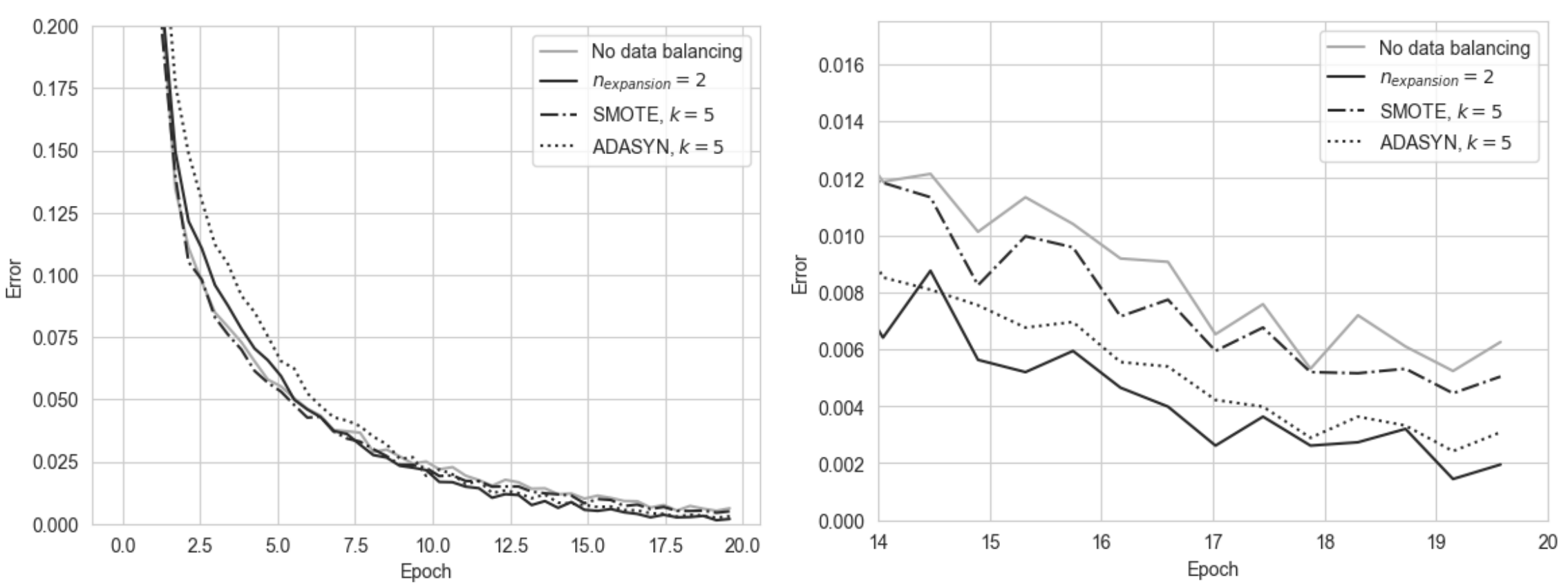

Figure 18. A comparison of errors among different sampling methods. (b) An enlarged plot showing the final epochs.